%% file: conference_101719.tex
\documentclass[conference]{IEEEtran}
\IEEEoverridecommandlockouts

\usepackage{amsmath,amssymb,amsfonts}
\usepackage{algorithmic}
\usepackage{graphicx}
\usepackage{textcomp}

\usepackage[svgnames]{xcolor}
\usepackage[hyphens]{url}
\usepackage{pifont}
\usepackage{amsmath,amssymb,amsfonts}
\usepackage{graphicx}
\usepackage{textcomp}
\usepackage{tikz}
\usepackage{soul}
\usepackage[table]{xcolor}
\usepackage{booktabs}
\usepackage{multirow}
\usepackage{makecell}
\usepackage{diagbox} 
\usepackage{enumitem}
\usepackage[ruled, lined, linesnumbered, commentsnumbered, longend]{algorithm2e}
\usepackage{cite}
\usepackage{tabularx}
\usepackage{adjustbox}

\usepackage{cite}
\usepackage{amsmath,amssymb,amsfonts}
\usepackage{algorithmic}
\usepackage{graphicx}
\usepackage{textcomp}
\usepackage{xcolor}
\def\BibTeX{{\rm B\kern-.05em{\sc i\kern-.025em b}\kern-.08em
    T\kern-.1667em\lower.7ex\hbox{E}\kern-.125emX}}
\begin{document}
\input{macros}

\title{{\name}: \underline{O}utlier-\underline{A}ware LUT-Ba\underline{s}ed GEMM w\underline{i}th Dual-\underline{S}ide Quantization for LLM Inference Acceleration

}


\author{
\IEEEauthorblockN{
Xueying Wu, Baijun Zhou, Zhihui Gao, Yuzhe Fu, Qilin Zheng, Yintao He$^{\dagger}$, Hai Li
}
\IEEEauthorblockA{
\textit{Duke University, Durham, NC, USA} \\
\{xueying.wu, baijun.zhou, zhihui.gao, yuzhe.fu, qilin.zheng, yintao.he, hai.li\}@duke.edu \\
$^{\dagger}$Corresponding author
}
}



\maketitle

\begin{abstract}
Large language models (LLMs) have demonstrated impressive capabilities across a wide range of applications, but demand substantial memory and compute resources during inference. Existing quantization methods expose a trade-off between efficiency and accuracy: weight-only quantization (WOQ) incurs costly dequantization overheads, while integer weight-and-activation quantization (INT-WAQ) reduces precision and degrades model quality. Non-uniform weight-and-activation quantization (NU-WAQ) can better capture the non-uniform distributions of LLM weights and activations, yet remains incompatible with conventional low-precision compute units.

This paper presents {\namebf}, a lookup table (LUT)-based architecture that enables efficient general matrix multiplication (GEMM) between non-uniformly quantized weights and activations without requiring dequantization. 
{\namebf} employs pre-computed Cartesian Product LUTs, achieving a 64$\times$ reduction in LUT size and enabling a 1024$\times$ higher computational parallelism over existing LUT-based GEMM methods.
To preserve accuracy under aggressive activation quantization, {\namebf} introduces an outlier-aware quantization scheme with concurrent LUT-based GEMM and error compensation for outliers.
Furthermore, we design \textit{Orizuru}, an efficient top-k detection engine for real-time activation outlier identification.

According to extensive evaluations, {\namebf} incurs an average accuracy drop of only 1.94\% compared to the FP16 baseline, which is 6.34\% lower than Atom.
On the hardware side, {\namebf} achieves an average 3.00$\times$ speedup and a 1.44$\times$ energy efficiency improvement compared to the FIGLUT accelerator.
\end{abstract}

\begin{IEEEkeywords}
large language models, non-uniform quantization, LUT-based computation, GEMM acceleration, hardware architecture, efficient inference
\end{IEEEkeywords}

\input{1_introduction}

\input{2_motivation}

\input{3_algorithm}

\input{4_hardware}
\input{5_evaluation}

\input{6_relatedwork}

\input{7_conclusion}

\newpage
\bibliographystyle{IEEEtranS}
\bibliography{reference}

\end{document}

%% file: macros.tex
\definecolor{lightgray}{gray}{0.9}
\definecolor{lightblue}{rgb}{0.9,0.9,1}
\definecolor{LightMagenta}{rgb}{1,0.5,1} 
\definecolor{red}{rgb}{1,0,0}
\definecolor{myBlueViolet}{rgb}{0.18,0.20,0.57}
\definecolor{myDarkBlue}{rgb}{0.25,0.32,0.49}

\newcommand\couldremove[1]{{\color{lightgray} #1}}
\newcommand{\remove}[1]{}
\newcommand{\move}[2]{ {\textcolor{Purple}{ \bf --- MOVE #1: --- }} {\textcolor{Orchid}{#2}} }

\newcommand{\hlc}[2][yellow]{ {\sethlcolor{#1} \hl{#2}} }
\newcommand\note[1]{\hlc[SkyBlue]{-- #1 --}} 

\newcommand\mynote[1]{\hlc[yellow]{#1}}

\newcommand{\xueying}[1]{{\textcolor{Orchid}{XW: {#1}}}}
\newcommand{\zhihui}[1]{{\textcolor{BurntOrange}{ZG: {#1}}}}
\newcommand{\baijun}[1]{{\textcolor{OliveGreen}{BZ: {#1}}}}
\newcommand{\fixme}[1]{{\textcolor{red}{{#1}}}}
\newcommand{\mymod}[1]{{\textcolor{Blue}{{#1}}}}

\newcommand\name{{\sc OASIS}}
\newcommand\namebf{{\sc\textbf{OASIS}}}

\newcommand{\myparatight}[1]{\vspace{0.5ex}\noindent\textbf{#1~~}}

\newcommand{\seqLen}{n}
\newcommand{\hiddenLen}{d}

\newcommand{\weightMat}{\mathbf{W}}
\newcommand{\weightVec}{\mathbf{w}}
\newcommand{\weight}{w}
\newcommand{\weightQuantMat}{\mathbf{W}_q}
\newcommand{\weightQuantVec}{\mathbf{w}_q}
\newcommand{\weightQuant}{w_Q}
\newcommand{\wiki}{WikiText-2}

\newcommand{\xMat}{\mathbf{X}}
\newcommand{\x}{x}
\newcommand{\xQuant}{x_q}

\newcommand{\KMeans}[1]{\textsf{KMeans}(#1)}

\newcommand{\cmark}{\textcolor{green!60!black}{\ding{51}}} 
\newcommand{\xmark}{\textcolor{red!80!black}{\ding{55}}}   
\definecolor{headergray}{RGB}{240,240,240}

\definecolor{iceblue}{RGB}{174,194,221}
\definecolor{lightgreen}{RGB}{169,209,158}
\definecolor{lightblue}{RGB}{217,236,236}
\definecolor{darkred}{RGB}{176, 36, 24}

\newcommand{\circlednumber}[3]{%
  \tikz[baseline=(char.base)]{
    \node[shape=circle,fill=#1,inner sep=2pt] (char) {#2\textcolor{black}{#3}};}} 
    
\newcommand\circlednumberblue[1]{%
  \begin{tikzpicture}[baseline=(char.base)]
    \node[shape=circle,,fill=iceblue,inner sep=1pt] (char) {\textcolor{black}{\scriptsize\sffamily\bfseries#1}};
  \end{tikzpicture}}

\newcommand\circlednumbergreen[1]{%
  \begin{tikzpicture}[baseline=(char.base)]
    \node[shape=circle,,fill=lightgreen,inner sep=1pt] (char) {\textcolor{black}{\scriptsize\sffamily\bfseries#1}};
  \end{tikzpicture}}

\newcommand\circlednumberbluetiny[1]{%
    \begin{tikzpicture}[baseline=(char.base)]
      \node[shape=circle,,fill=iceblue,inner sep=0pt] (char) {\textcolor{black}{\scriptsize\sffamily\bfseries#1}};
    \end{tikzpicture}}
    
\newcommand\circlednumbergreentiny[1]{%
    \begin{tikzpicture}[baseline=(char.base)]
      \node[shape=circle,,fill=lightgreen,inner sep=0pt] (char) {\textcolor{black}{\scriptsize\sffamily\bfseries#1}};
    \end{tikzpicture}}

\newcommand\circlednumberred[1]{%
  \tikz[baseline=(char.base)]{
    \node[
      shape=circle,
      fill=darkred,
      inner sep=0.3pt,
      minimum size=8pt,
      line width=0.3pt
    ] (char) {\fontsize{6.5}{6.5}\selectfont\bfseries\sffamily\textcolor{white}{#1}};
  }%
}

\newcommand\circlednumberblack[1]{%
  \tikz[baseline=(char.base)]{
    \node[
      shape=circle,
      fill=black,
      inner sep=0.3pt,
      minimum size=8pt,
      line width=0.3pt
    ] (char) {\fontsize{6.5}{6.5}\selectfont\bfseries\sffamily\textcolor{white}{#1}};
  }%
}

\newcommand\circlednumberwhite[1]{%
  \tikz[baseline=(char.base)]{
    \node[
      shape=circle,
      draw,
      inner sep=0.3pt,
      minimum size=8pt,
      line width=0.3pt
    ] (char) {\fontsize{6.5}{6.5}\selectfont #1};
  }%
}

%% file: 1_introduction.tex
\section{Introduction} \label{sec:introduction}

Large language models (LLMs) have achieved strong performance across diverse domains such as dialogue systems~\cite{achiam2023gpt,guo2025deepseek}, code generation~\cite{roziere2023code,tao2024crystal}, and electronic design automation~\cite{he2023chateda,fu2023gpt4aigchip}.
However, the rapidly increasing model sizes introduce substantial memory and computational costs during inference~\cite{kwon2023efficient,brown2020language}, motivating extensive research on model compression.

\begin{figure}[t]
    \centering
    \includegraphics[width=1\linewidth]{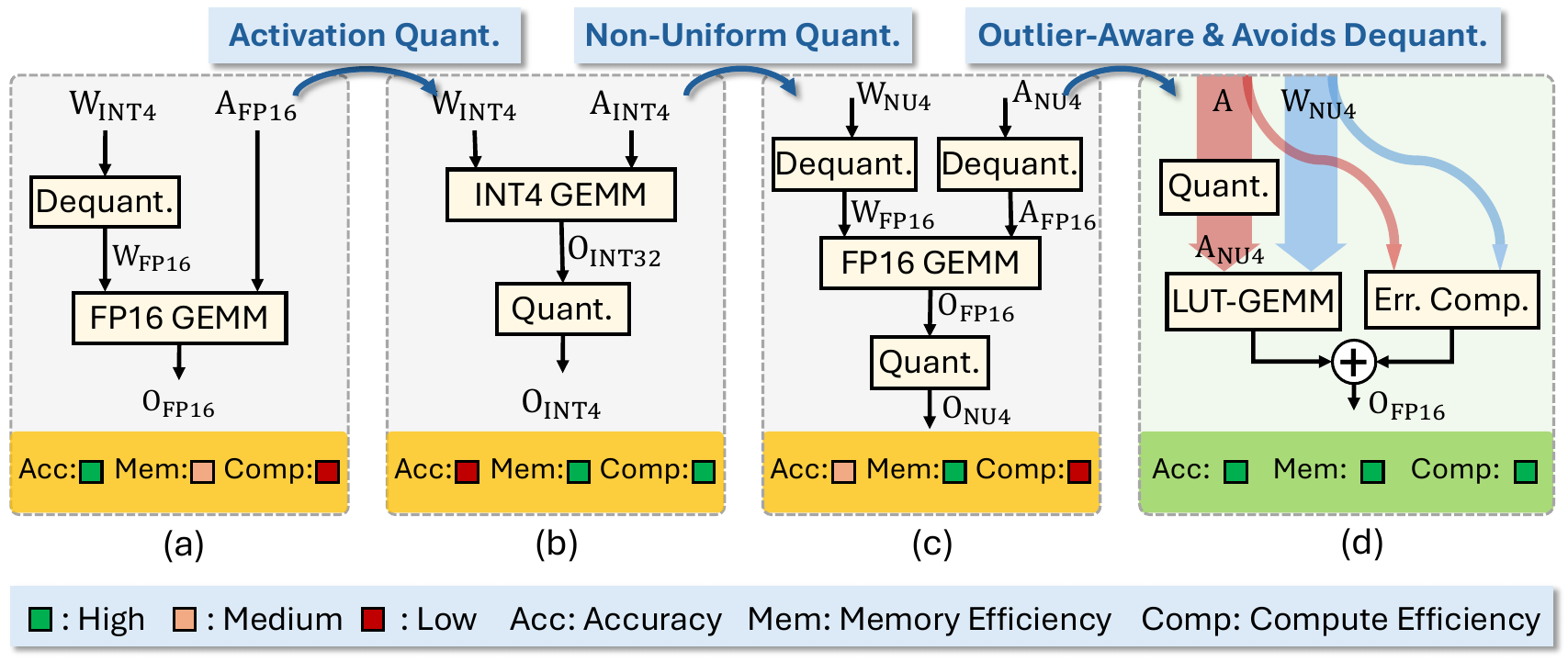}
    \vspace{-0.3in}
    \caption{
    Comparison of GEMM schemes:
    (a) $\text{W}_{\text{INT4}}\text{A}_{\text{FP16}}$ FP16 GEMM,
    (b) $\text{W}_{\text{INT4}}\text{A}_{\text{INT4}}$ INT4 GEMM,
    (c) $\text{W}_{\text{NU4}}\text{A}_{\text{NU4}}$ FP16 GEMM,
    (d) $\text{W}_{\text{NU4}}\text{A}_{\text{NU4}}$ LUT-GEMM (Ours).
    ``NU” refers to non-uniform quantization.
    Our proposed design features two parallel computation branches that separately handle inlier and outlier operations.
    Our LUT-based GEMM design enables matrix multiplications between non-uniformly quantized weights and activations without requiring dequantization.
    }
    \vspace{-5mm}
    \label{fig:gemm_compare}
\end{figure}

Among these efforts, weight-only quantization (WOQ)~\cite{frantar2022gptq,lin2024awq} stands out for effectively reducing memory footprint while preserving model accuracy. Yet WOQ still suffers from a key limitation: weights and activations reside in different numerical formats. 
Consequently, WOQ necessitates dequantizing weights to FP16 before executing, which is shown in Fig.~\ref{fig:gemm_compare}(a). This dequantization step can dominate GEMM execution time, often constituting 20-90\% of the total runtime~\cite{lin2024qserve,li2023fast}.
To eliminate the dequantization cost, recent studies~\cite{mo2025lut,park2022lut,park2025figlut} have proposed LUT-based GEMM methods that utilize lookup tables (LUTs) to directly execute GEMMs between quantized weight and FP16 activations. 
However, their performance remains constrained by the overhead of on-the-fly LUT generation, large LUT sizes and limited parallelism.

Weight-and-activation quantization (WAQ) offers a more promising approach by enabling fully low-precision GEMMs, reducing memory footprint for both weights and KV-cache memory, and fundamentally removing the need for mixed-format computation~\cite{ashkboos2024quarot,liu2024spinquant,gao2026disaggregated}.
However, existing WAQ methods exhibit a fundamental trade-off between accuracy and compute efficiency.
As shown in Fig.~\ref{fig:gemm_compare}(b), integer WAQ (INT-WAQ) quantizes both operands into low-bit integers that existing low-precision hardware can directly process, but its limited representation capability often causes substantial accuracy degradation~\cite{xiao2023smoothquant,zhao2024atom}.
In contrast, non-uniform WAQ (NU-WAQ), especially learned-codebook methods~\cite{kim2023squeezellm,hooper2024kvquant}, represents each value using an index that selects from a learned set of codebook entries, allowing the quantized values to more accurately follow the underlying data distribution.
NU-WAQ significantly improves accuracy, but its index-coded data format is incompatible with existing low-precision compute units~\cite{nvidia_turing_whitepaper_2018,nvidia-a100}.
As a result, NU-WAQ execution on existing hardware must dequantize values back to FP16 before executing GEMMs, which is illustrated in Fig.~\ref{fig:gemm_compare}(c).
This negates the computational advantages of quantization and yields poor computational efficiency.

To resolve the dilemma between \textit{high-efficiency but low-accuracy INT-WAQ} and \textit{high-accuracy but low-efficiency NU-WAQ} (Fig.~\ref{fig:gemm_compare}), we propose a LUT-based method that enables efficient GEMMs between non-uniformly quantized weights and activations without requiring dequantization.
Prior LUT-based GEMM methods were developed to optimize GEMM computations with WOQ~\cite{mo2025lut,park2022lut,park2025figlut}. 
They rely on large inner-product LUTs that must be regenerated for dynamic activations, leading to excessive LUT size and limited parallelism. 
On the other hand, we observe that WAQ fundamentally unlocks three key opportunities to achieve a far more efficient LUT-based GEMM design:
(1) In learned-codebook WAQ, both weight and activation centroids are learned offline, allowing the entire LUT to be pre-computed before inference and eliminating on-the-fly LUT generation.
(2) Because both operands are quantized, the space of possible multiplication outcomes is greatly reduced, enabling us to store Cartesian Product entries instead of full inner products, which substantially shrinks LUT size.
(3) The Cartesian Product LUT is independent of reduction length, enabling larger compute granularity and significantly higher parallelism during GEMMs.

To further retain model accuracy in WAQ, it is important to carefully handle the activation outliers.
This is because activation outliers exhibit both higher quantity and magnitude compared to weight outliers, resulting in high quantization noises~\cite{ashkboos2024quarot, xiao2023smoothquant,lin2024awq}.
Therefore, we propose an outlier-aware mechanism that identifies activation outliers during inference and compensates for their errors without incurring additional runtime overhead.

Incorporating the aforementioned optimizations, we propose {\name}: an \underline{O}utlier-\underline{A}ware LUT-Ba\underline{s}ed GEMM Scheme w\underline{i}th Dual-\underline{S}ide Quantization for LLM Inference Acceleration.
We compare the end-to-end LLM inference performance of {\name} with state-of-the-art (SOTA) quantization algorithms and accelerators.
For the algorithmic performance, on average, {\name} achieves an accuracy degradation of only 1.94\% compared to the FP16 baseline.
This accuracy degradation is 6.34\% lower than Atom~\cite{zhao2024atom}.
For the hardware performance, {\name} achieves average speedups of 3.00$\times$ and energy efficiency improvements of 1.44$\times$ over FIGLUT~\cite{park2025figlut}.
The main contributions of this work include:
\begin{itemize}[leftmargin=1em]
\item We propose WAQ LUT-GEMM, a LUT-based GEMM method enabling efficient computation between non-uniformly quantized weights and activations without dequantization. It uses pre-computed Cartesian Product LUTs to significantly reduce LUT size and improve parallelism.
\item We introduce an outlier-aware quantization scheme with look-ahead computation and error compensation to efficiently handle activation outliers during inference.
\item We propose the architecture design of the {\name} accelerator, which efficiently supports WAQ LUT-GEMM.
\item We develop \textit{Orizuru}, a lightweight top-$k$ engine for identifying activation outliers in real-time data streams.
\end{itemize}

%% file: 2_motivation.tex
\section{Backgrounds and Motivations} \label{sec:motivation}
\subsection{LLM Quantization}

LLMs typically utilize FP16 format for weights and activations during inference~\cite{dubey2024llama,achiam2023gpt}.
Quantization, which reduces the numerical precision of weights and activations, has emerged as a promising technique to optimize LLM inference efficiency~\cite{frantar2022gptq,kim2023squeezellm,lin2024awq,hooper2024kvquant,guo2025survey}.
WOQ methods effectively reduce model memory footprint by quantizing the weights~\cite{frantar2022gptq,lin2024awq}, but fail to leverage emerging efficient low-precision compute units (e.g., GPUs' INT4 Tensor Cores~\cite{nvidia_turing_whitepaper_2018}) since the activations remain in FP16 format.
In contrast, WAQ methods quantize both weights and activations, enabling low-precision computation for faster inference~\cite{ashkboos2024quarot,liu2024spinquant}.
Therefore, recent studies have increasingly focused on developing WAQ methods to further enhance LLM inference efficiency~\cite{zhao2024atom, lin2025duquant, sun2024flatquant, liu2024spinquant, ashkboos2024quarot}.

Conventional WAQ methods quantize both weights and activations into integer representations, which face significant accuracy degradation in low-precision quantization configurations~\cite{ashkboos2024quarot,zhao2024atom,liu2024spinquant,10547183}.
Compared to integer quantization, which maps floating-point numbers to equally spaced integer levels, non-uniform quantization employs unequally spaced centroids that better align with the actual distributions of weights and activations, thereby reducing quantization noise~\cite{zhang2018lq,macqueen1967some,tseng2025training,alvarez2025introducingNVFP4}.
To achieve higher accuracy, recent studies have leveraged non-uniform quantization methods to achieve low-precision LLMs~\cite{kim2023squeezellm, liu2023llm, park2025figlut}.

Non-uniform quantization includes low-precision floating-point formats (e.g., MXFP4~\cite{tseng2025training}, NVFP4~\cite{alvarez2025introducingNVFP4}) and learned-codebook methods~\cite{zhang2018lq,macqueen1967some}. 
The learned-codebook methods, which optimize centroids via training, generally achieve higher accuracy~\cite{kim2023squeezellm, hooper2024kvquant}. 
These methods represent data as integer index matrices that reference a floating-point codebook. 
For example, K-Means quantization~\cite{macqueen1967some} can be written as:
\begin{equation}
    \tilde{x}_i = C_{idx_i},\qquad idx_i = \arg\min_k \| x_i - C_k \|^2,
\end{equation}
where \(x_i\) is the original data point, \(\tilde{x}_i\) is its quantized value, \(C_k\) is the centroid of the $k^{th}$ cluster, and \(idx_i\) is the integer index corresponding to \(\tilde{x}_i\) that selects the nearest centroid.
To represent a matrix of size \(M \times N\) with \(n\)-bit K-Means quantization, an \(n\)-bit integer index matrix \(idx\) of size \(M \times N\) and an FP centroid codebook \(C\) of size \(2^n\) are required.

Although learned-codebook schemes reduce quantization error, they incur substantial runtime cost at inference because current accelerators do not natively support such non-uniform data representations~\cite{nvidia-a100,nvidia2025blackwell}. 
Performing GEMMs with index-coded data therefore requires per-element codebook lookups and dequantization into FP16 values, followed by FP16 GEMMs, which adds significant overhead~\cite{kim2023squeezellm,hooper2024kvquant}. Therefore, unleashing the performance benefits of learned-codebook quantization motivates the design of hardware that directly supports their non-uniform data representations.

\subsection{LUT-Based GEMM Schemes} \label{sec:LUT_compare}

\begin{figure}
    \centering
    \includegraphics[width=1\linewidth]{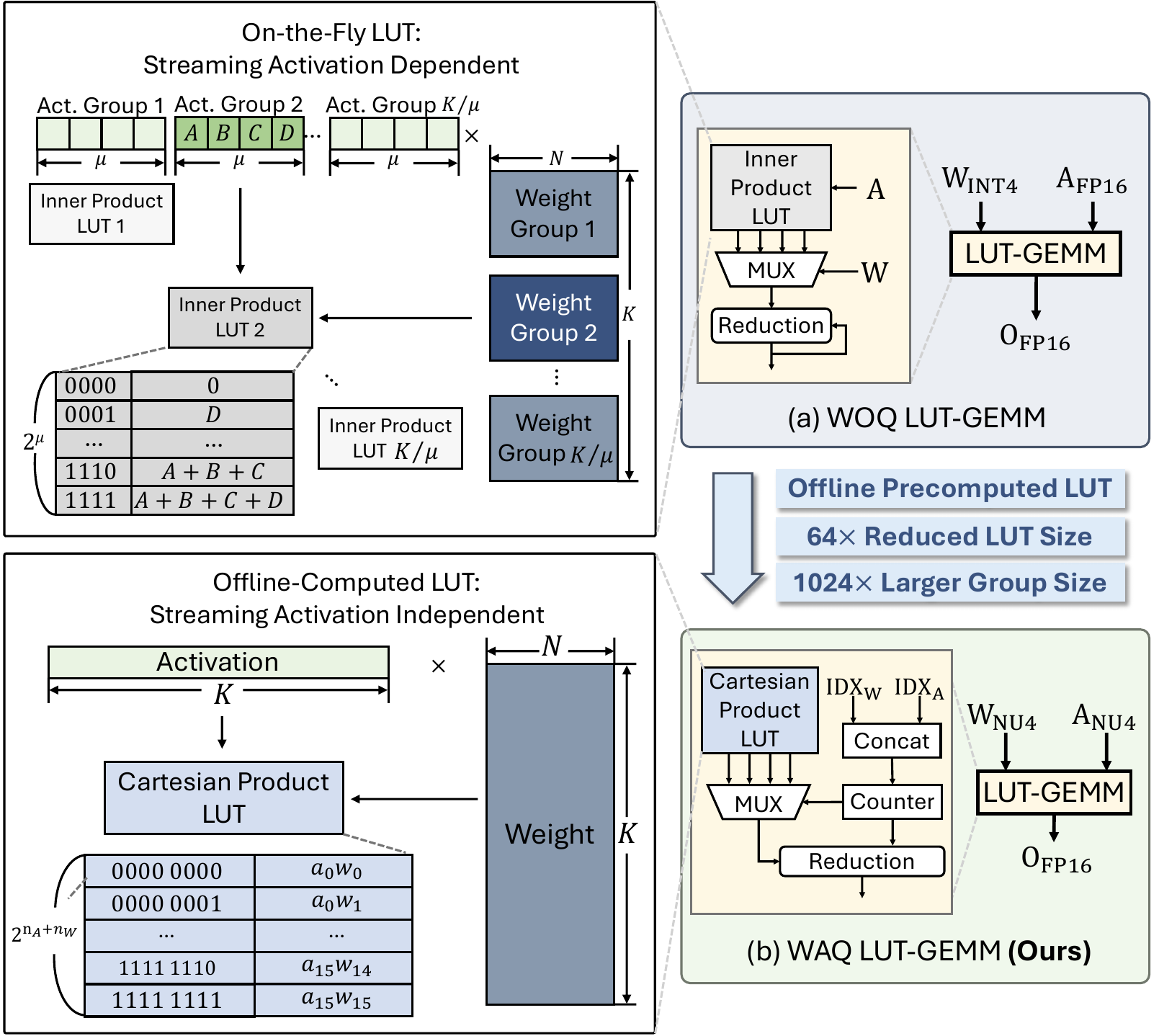}
    \vspace{-6mm}
    \caption{(a) Existing WOQ LUT-GEMM scheme. $A$, $B$, $C$, and $D$ denote different streaming activation values.
    (b) Our proposed WAQ LUT-GEMM scheme. $a_i$ and $w_i$ denote the activation and weight centroids, respectively.
    }
    \vspace{-3mm}
    \label{fig:lut_compare}
\end{figure}

\renewcommand{\arraystretch}{1.3}
\begin{table}
\centering
\small
\caption{Comparison of LUT-based GEMM schemes.}
\begin{tabularx}{1.0\linewidth}{l|c|c}
\toprule
\rowcolor{headergray}
& \textbf{WOQ LUT-GEMM} & \textbf{Ours} \\
\midrule
Wgt. Precision ($n_W$) & NU4 & NU4 \\
\hline
Act. Precision ($n_A$) & FP16 & NU4 \\
\hline
Offline-Computed LUT? & \xmark & \cmark \\
\hline
Group Size & $\mu$ & $K$ \\
\hline
LUT Size & $2^{\mu}\cdot\frac{K}{\mu}$ & $2^{n_A+n_W}$ \\
\hline
\#FLOPs for Reduction & $\frac{K}{\mu}\cdot n_W \cdot N$ & $2^{n_A+n_W} \cdot N$ \\
\bottomrule
\end{tabularx}
\label{tab:lut_gemm_compare}
\end{table}

LUT-based computation has been demonstrated to be an effective approach for performing efficient GEMMs without dequantization cost~\cite{park2025figlut,park2022lut,mo2025lut}.
Generally, LUT-based GEMM methods take two inputs: which kind of multiplication result is stored in the LUT, and how the LUT is indexed.
As shown in Fig.~\ref{fig:lut_compare}(a) and Table~\ref{tab:lut_gemm_compare}, existing LUT-GEMM methods for WOQ execution~\cite{park2025figlut,park2022lut,mo2025lut} store the group-wise inner product results between weights and activations in the LUT and use the weights as the MUX select signals.
However, they remain compute-inefficient due to the following limitations:
(1) The LUT depends on the streaming activations and therefore must be generated on-the-fly;
(2) Let $K$ denote the reduction length of the inner product, a LUT with the size of $2^K$ is required to store all possible inner product results between weights and FP16 activations, which is impractical for large reduction lengths (e.g., \(K = 4096\) in LLaMA-7B~\cite{touvron2023llama}).
To limit the LUT size, existing LUT-based methods partition the weights and activations into small groups of size \(\mu\) and perform multiple partial inner products.
(3) Partial-sum reductions are required across multiple groups, incurring additional FLOPs and latency.
Existing WOQ LUT-GEMM methods adopt bit-serial weight processing to further reduce the LUT size, with each cycle only processing one bit of the weights~\cite{park2025figlut,park2022lut,mo2025lut}.

To balance between the LUT size and computational parallelism, weights and activations are grouped into groups of size \(\mu = 4\)~\cite{park2025figlut,mo2025lut}.
Among the existing WOQ LUT-GEMM methods, FIGLUT~\cite{park2025figlut} and LUT Tensor Core~\cite{mo2025lut} manage to reduce the LUT size by half by using the most-significant-bit (MSB) of the weight index as the enable signal of the negation logic.
However, the LUT sizes of these methods are still large, and the computational parallelism remains limited.

While WOQ LUT-GEMM methods can be adapted to avoid dequantizing weights and activations in NU-WAQ GEMMs, we observe three opportunities specific to NU-WAQ that further improve the compute efficiency:
(1) In NU-WAQ, activation centroids are trained offline, so the set of possible activation values at runtime is known in advance, allowing the LUT to be precomputed and stored on-chip prior to inference.
(2) With both weights and activations quantized, the number of distinct Cartesian Product results is limited; storing the Cartesian Product of weight and activation centroids in the LUT instead of the inner product results substantially reduces LUT sizes.
(3) A Cartesian Product LUT is independent of the reduction length, allowing computation at a larger granularity and significantly eliminating the number of FLOPs during reductions.

Leveraging the aforementioned opportunities, we propose a novel LUT-based GEMM scheme for NU-WAQ, as illustrated in Fig.~\ref{fig:lut_compare}(b).
The inputs of our design are weights and activations that are non-uniformly quantized using learned codebooks.
The LUT stores the Cartesian Product of weight and activation centroids, which can be precomputed and loaded on-chip before inference. 
At runtime, weight and activation indices are concatenated.
The occurrence counts of the unique concatenated indices are calculated and used to perform reductions as weighted sums of the corresponding LUT entries. 
These unique concatenated indices act as MUX select signals to fetch the corresponding Cartesian Product values from the LUT.

The configuration comparison between existing WOQ LUT-GEMM methods and our proposed WAQ LUT-GEMM method is summarized in Table~\ref{tab:lut_gemm_compare}.
$n_W$ and $n_A$ denote the bit-widths of weights and activations, respectively.
Consider an $M-K-N$ GEMM example between weights and activations where $M = 1$ and $N = K = 4096$, which is a common case in the LLaMA-7B model~\cite{touvron2023llama}.
For the configuration of $n_W = n_A = 4$, our proposed WAQ LUT-GEMM method achieves a 64$\times$ reduction in LUT size,
a 1024$\times$ increase in group size, and a 16$\times$ reduction in floating-point operations (FLOPs) for reductions over existing WOQ LUT-GEMM methods~\cite{mo2025lut,park2025figlut,park2022lut}.

\subsection{Handling Activation Outliers in WAQ}    \label{sec:activation_outlier_bg}

\begin{figure}[t]
    \centering
    \includegraphics[width=1\linewidth]{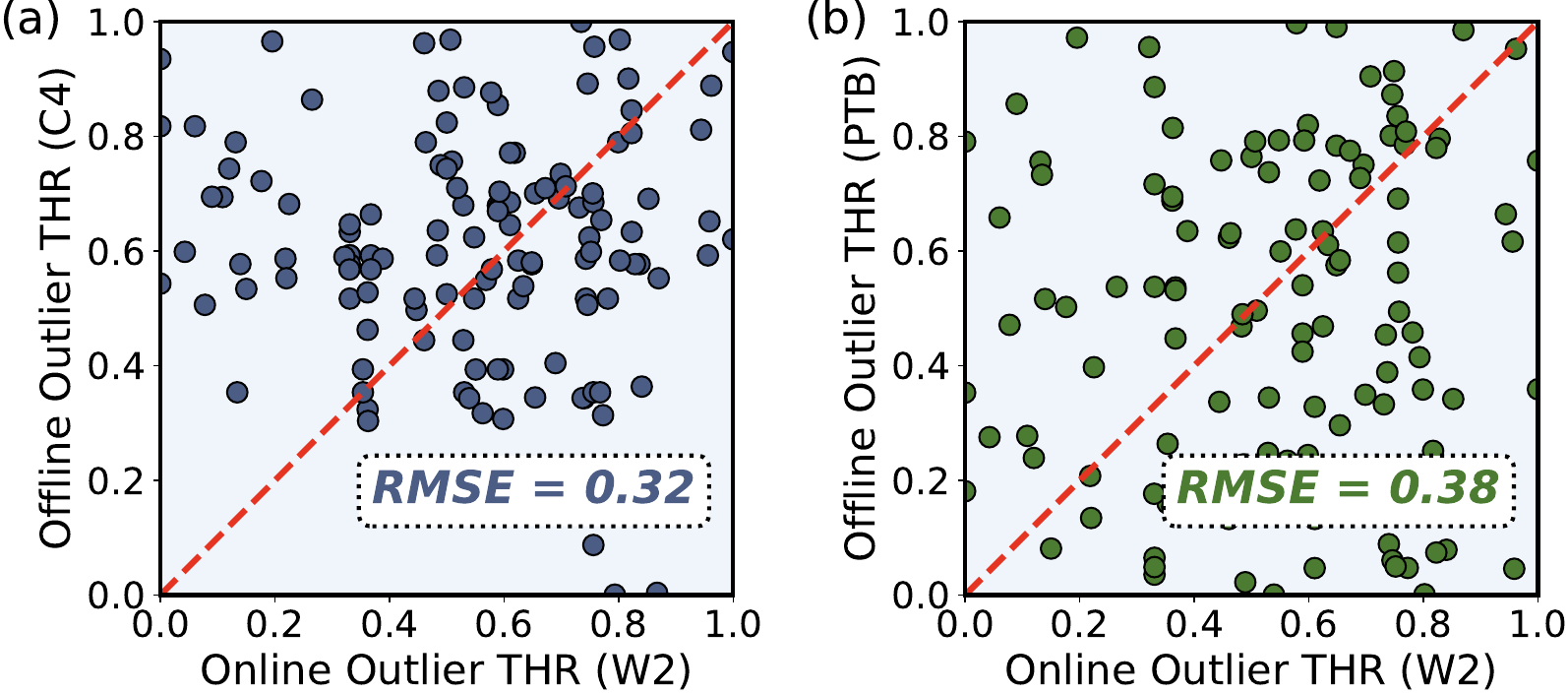}
    \vspace{-0.3in}
    \caption{
    Comparison between online and offline derived upper activation outlier thresholds.
    The online dataset is WikiText-2~\cite{merity2016pointer} (W2), and the offline dataset is (a) C4~\cite{dodge2021documenting} and (b) PTB~\cite{marcus1994penn}.
    The activations used to compute the thresholds and centroids are the input of the $1^{st}$ $q\_proj$ layer in the LLaMA-3-8B model~\cite{grattafiori2024llama}.
    128 activation tokens are applied to compute the thresholds, each with the dimension of $1 \times 4096$.
    The thresholds are normalized to [0, 1].
    }

    \label{fig:thr}
\end{figure}

\begin{figure}
    \centering
    \includegraphics[width=1\linewidth]{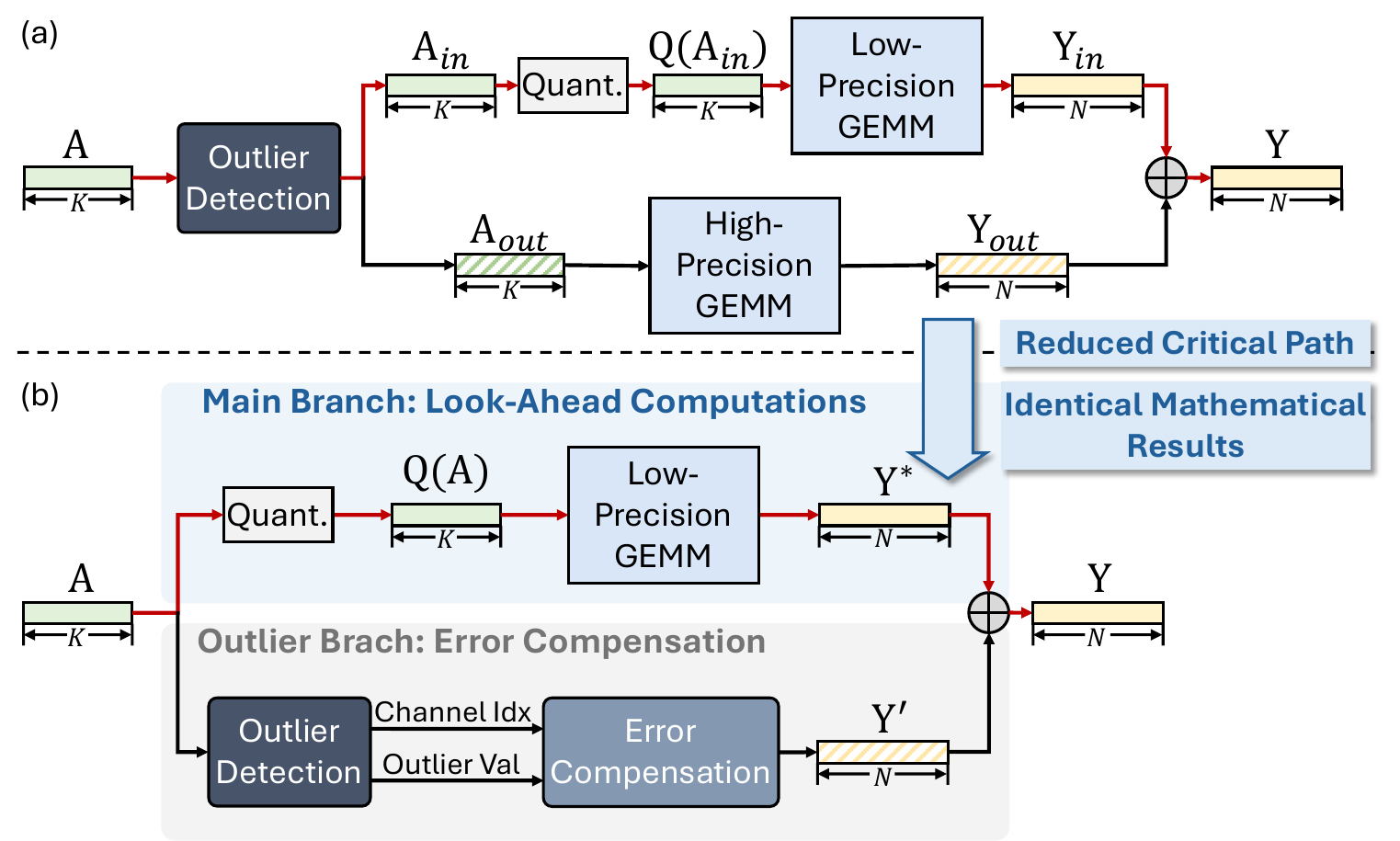}
    \vspace{-6mm}
    \caption{
    Comparison of (a) conventional dynamic outlier detection and (b) the proposed look-ahead scheme.  }
    \vspace{-3mm}
    \label{fig:look_ahead}
\end{figure}

In LLMs, activation outliers exhibit both higher quantity and magnitude compared to weight outliers, which expands the quantization range and reduces the effective bit resolution available for the majority of values (inliers)~\cite{liu2024spinquant}.
To mitigate activation outlier-induced quantization errors, prior works have proposed to isolate outliers and preserve them in higher-precision representations (e.g., INT8 or FP16) during computation~\cite{zhao2024atom,hooper2024kvquant}.
Existing works basically identify outliers with two approaches:
(1) dynamically identify top-k outliers during inference~\cite{hooper2024kvquant}; and
(2) leveraging an offline calibration dataset to identify certain activation channels that contain a large volume of outliers as outlier channels. 
During online inference, the channel indices are used to indicate which activation values are outliers~\cite{zhao2024atom,liu2025micromix}.
It has been evaluated that dynamic outlier detection generally yields higher accuracy than static outlier channel identification~\cite{hooper2024kvquant}.
To explain this, we define the value of the top-k largest activation as the upper outlier threshold. 

Fig.~\ref{fig:thr} demonstrates substantial discrepancies in upper outlier thresholds between offline and online datasets.
The online calibration uses WikiText-2~\cite{merity2016pointer}, while online inference uses C4~\cite{dodge2021documenting} and PTB~\cite{marcus1994penn} in Fig.~\ref{fig:thr}(a) and (b), respectively.
The root mean square error (RMSE) between online and offline thresholds is 0.32 and 0.38 in Fig.~\ref{fig:thr}(a) and (b), respectively, indicating low similarity.

Therefore, to preserve model accuracy, it is crucial to dynamically identify outliers during inference.

Fig.~\ref{fig:look_ahead}(a) illustrates the conventional workflow of dynamic outlier detection during inference~\cite{hooper2024kvquant}.
Generally, the entire activation vector is scanned to identify outliers, separating the activations into two groups: inliers ($A_{in}$), which are quantized into low-precision formats; and outliers ($A_{out}$), which are retained in higher-precision formats.
The activation inliers and outliers undergo low-precision GEMMs and high-precision GEMMs with the weights, respectively, and their results $Y_{in}$ and $Y_{out}$ are combined to produce the final output.
While the GEMMs associated with inliers and outliers can be executed in parallel on separate compute units, the outlier detection process stays on the \textbf{\textit{critical path}} (indicated with red arrows), incurring significant latency overhead.

To address this issue, we propose a look-ahead computation scheme, as illustrated in Fig.~\ref{fig:look_ahead}(b).
To avoid introducing additional runtime overhead for handling outliers, we propose two concurrent compute branches:
the \textit{main branch} performs \textbf{\textit{look-ahead}} WAQ-LUT GEMMs with the entire activation quantized, ignoring the quantization errors from outliers;
meanwhile, the \textit{outlier branch} calculates the quantization errors from outliers and performs \textbf{\textit{error compensation}}.
The final GEMM result is obtained by summing the outputs of look-ahead GEMMs ($Y^*$) and error compensation ($Y'$), which leads to identical mathematical results as conventional dynamic outlier detection, but without incurring additional latency.

\begin{figure}[t]
    \centering
    \includegraphics[width=1\linewidth]{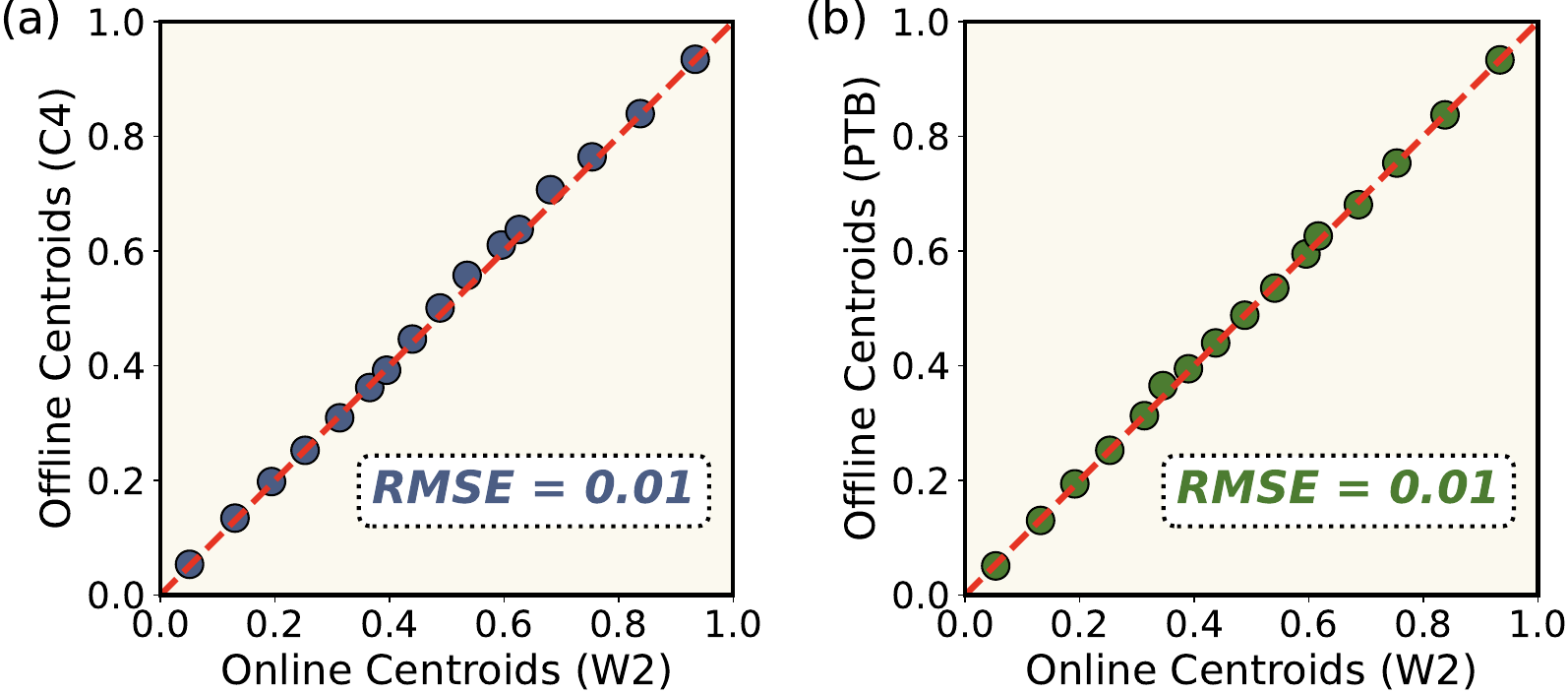}
    \vspace{-0.3in}
    \caption{
    Comparison between online and offline derived 4-bit quantization centroids for activations.
    The configurations of activations are the same as those in Fig.~\ref{fig:thr}.
    The centroids are normalized to [0, 1].
    }
    \label{fig:centroids}
\end{figure}

\subsection{Our Approach}
We aim to achieve efficient LLM inference with high model performance by leveraging NU-WAQ.
Driven by the aforementioned observations and analysis, in this paper, we propose {\name}, an algorithm-architecture co-design framework.
Specifically, \S~\ref{sec:algorithm} details the computation optimizations of {\name}, which focus on the following two aspects: 
(1) to enable efficient GEMMs between the inliers of non-uniformly quantized weights and activations, we propose a WAQ LUT-GEMM scheme without requiring dequantization; and
(2) we introduce an outlier-aware GEMM computation scheme that concurrently performs computations on both activation inliers and outliers.
We also present the architecture design of the {\name} accelerator in \S~\ref{sec:hardware}, which efficiently implements the proposed computation schemes.

%% file: 3_algorithm.tex
\section{Computation Optimizations}\label{sec:algorithm}
In this section, we present the computation optimizations to enable efficient LLM inference with NU-WAQ.
Specifically, \S~\ref{sec:quant} describes the quantization method used in {\name},
\S~\ref{sec:WAQ_LUT} presents the WAQ LUT-based GEMM scheme for efficient GEMMs with non-uniformly quantized weights and activations,
and \S~\ref{sec:look_ahead} introduces the look-ahead computation and error compensation designs to handle activation outliers.

\subsection{Quantizing Weights and Activations} \label{sec:quant}

To maintain model performance in low-precision configurations, {\name} adopts the learned-codebook quantization method of K-Means~\cite{macqueen1967some} for weight and activation quantization.
Specifically, we employ output-channel-wise quantization for weights and token-wise quantization for activations.
For weights, the entire weight matrix shares the same quantization centroids, while each output channel has its own scaling factor.
For activations, each token has its own set of quantization centroids and scaling factors.
We quantize the pretrained LLM weights to obtain the weight centroids, while the activation centroids are learned through an offline calibration dataset.
As shown in Fig.~\ref{fig:centroids}, the offline and online activation centroids exhibit high consistency across different dataset configurations.
The RMSE values between the offline and online centroids are both only 0.01 in Fig.~\ref{fig:centroids}(a) and (b).
This indicates the feasibility of using offline-learned activation centroids for online activation quantization to avoid the overhead of activation centroid learning during inference.
To further mitigate model performance degradation caused by activation quantization, we dynamically identify the top-0.5\% largest and the bottom-0.5\% smallest of the activations as outliers and preserve them in FP16 format.
\subsection{WAQ LUT-Based GEMM} \label{sec:WAQ_LUT}

\begin{figure}[t]
    \centering
    \includegraphics[width=\linewidth]{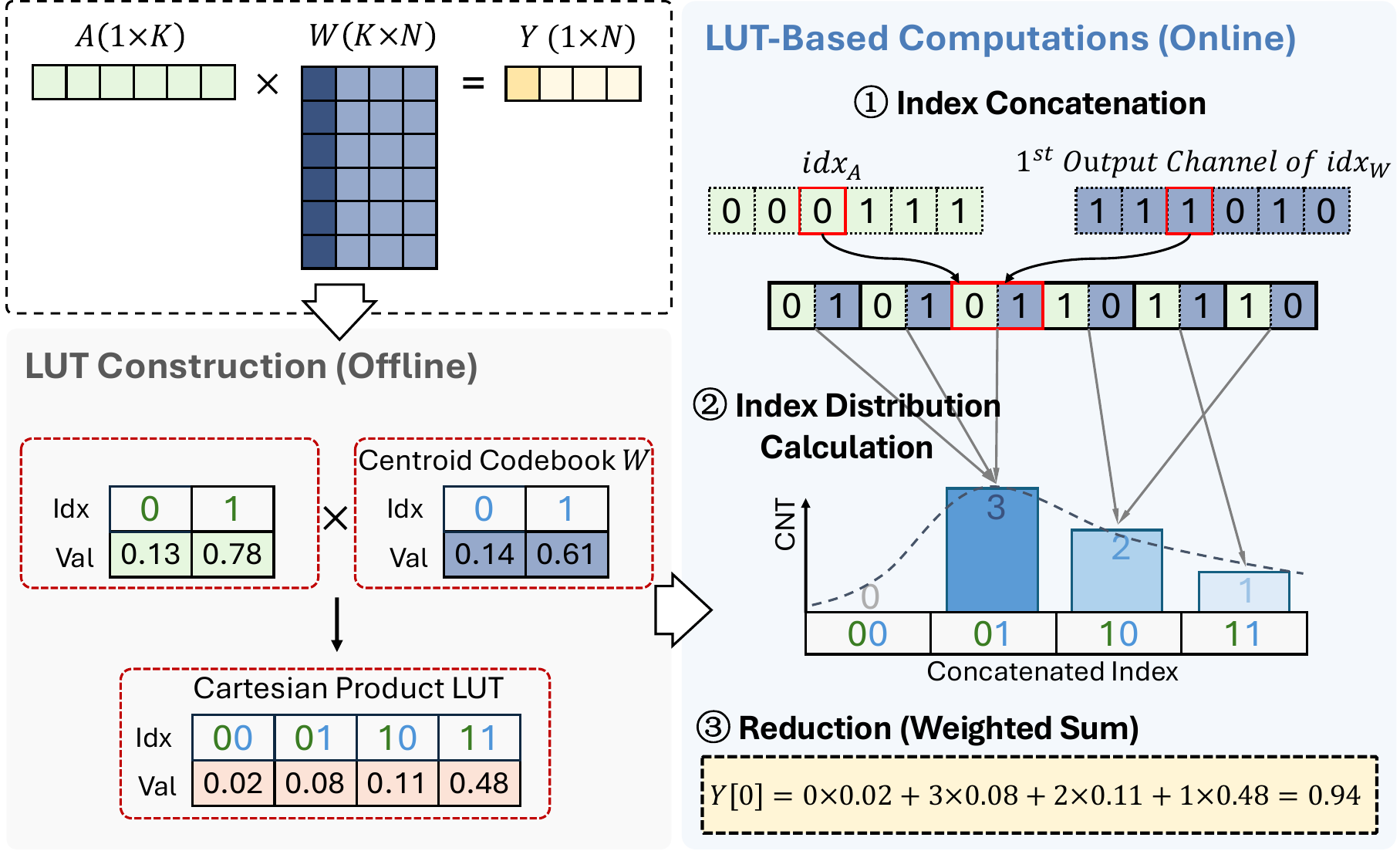}
    \vspace{-0.2in}
    \caption{WAQ LUT-based GEMM computation scheme.}
    \label{fig:WAQ_LUT_GEMM}
    \vspace{-0.1in}
\end{figure}

\begin{figure}[t]
    \centering
    \includegraphics[width=1\linewidth]{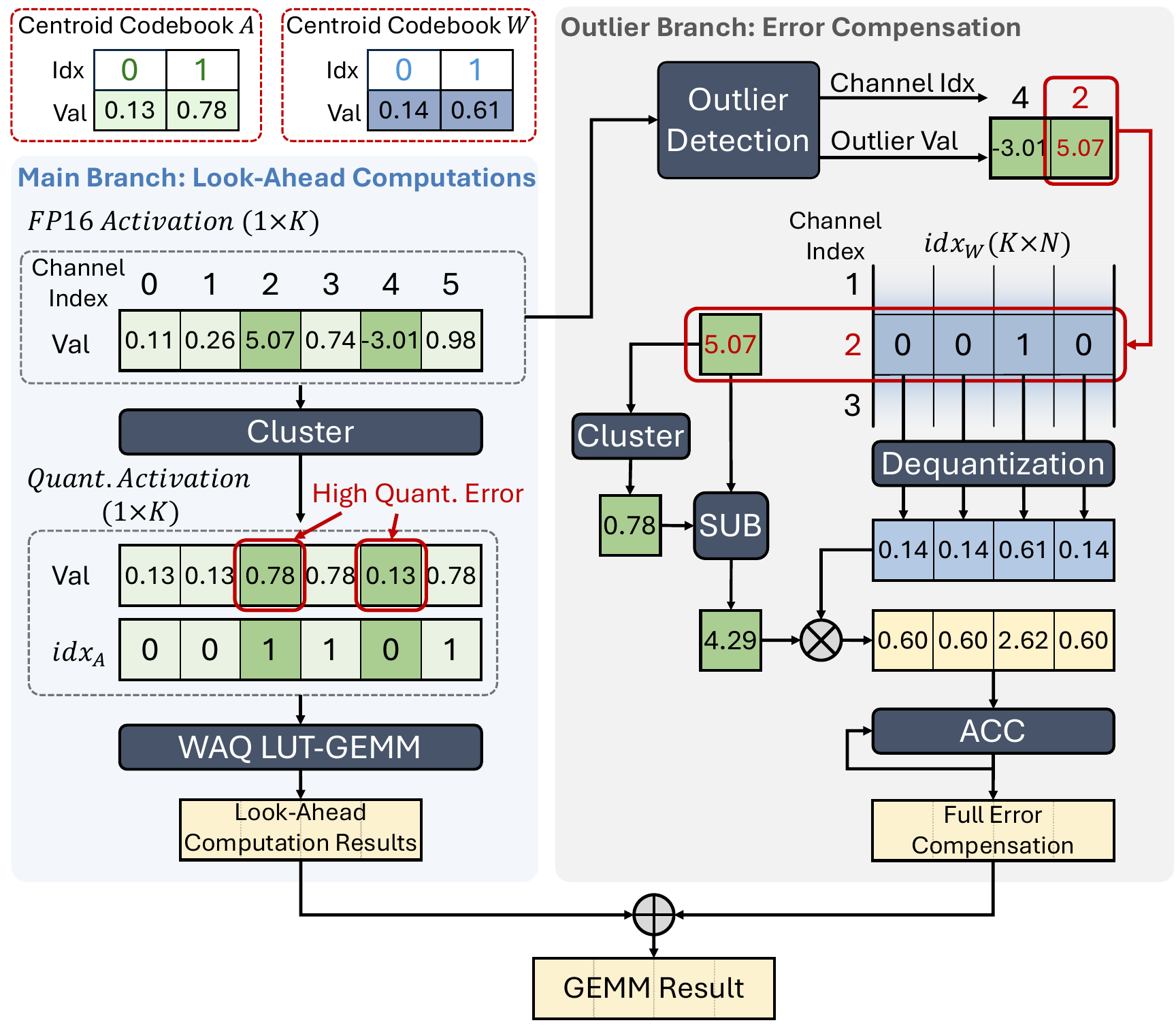}
    \vspace{-0.3in}
    \caption{Look-ahead computations and error compensation.}
    \vspace{-0.2in}
    \label{fig:error_compensation}
\end{figure}

Leveraging the WAQ-specific opportunities discussed in Section~\ref{sec:LUT_compare}, we propose a WAQ LUT-based GEMM scheme to efficiently execute GEMMs between non-uniformly quantized weights and activations.
Fig.~\ref{fig:WAQ_LUT_GEMM} shows an example of computing the GEMM output with the activation and the first output channel of weights using the proposed WAQ LUT-based GEMM scheme.
In this $M-K-N$ GEMM example, $M = 1, K = 6, N = 4$, and $n_W = n_A = 1$.

In learned-codebook WAQ methods, both the quantization centroids of weights and activations are determined offline. 
In other words, unique weight and activation values are predefined before inference. 
Therefore, we construct the Cartesian Product LUT offline, which stores all possible multiplication results between the weight and activation centroids.
During online inference, we concatenate the indices of the activations and weights, which is shown in step \circlednumberwhite{1}.
Then, in step \circlednumberwhite{2}, we calculate the distribution of these concatenated indices.
Finally, in step \circlednumberwhite{3}, we perform the reduction of the multiplication results along the input channel dimension ($K$).
Specifically, we replace the $K$ FP16 additions in the conventional GEMM with a weighted sum of the multiplication results stored in the Cartesian Product LUT.
The counts of each unique concatenated index serve as weights of the weighted sum.
The number of FP16 additions is reduced from $K$ to $2^{n_W + n_A}$, which is significantly smaller in low-precision quantization scenarios.

Consider the common case of W4A4 GEMMs, where $n_W = n_A = 4$.
The Cartesian product LUT stores $2^{n_W + n_A}$ multiplication results—only 256 entries.
This LUT size is 64$\times$ smaller than the inner-product LUT used in existing WOQ LUT-GEMM methods for a 4096$\times$4096 weight layer.
Therefore, unlike WOQ LUT-GEMM methods that require small group sizes to control LUT size, our design can support arbitrary reduction lengths without increasing LUT size, enabling higher parallelism.
In our WAQ LUT-GEMM scheme, the reduction length is equal to the input channel number of the weights $K$.
The high computational parallelism of our design yields a 16$\times$ reduction in FLOPs compared to existing WOQ LUT-GEMM methods.
These advantages become more pronounced for larger LLMs as per-layer input channel numbers increase~\cite{dubey2024llama}, which is evaluated in \S\ref{sec:ablation}.

\subsection{Look-Ahead Computations and Error Compensation} \label{sec:look_ahead}
As discussed in \S\ref{sec:activation_outlier_bg}, it is important to effectively hide the latency of outlier detection.
To address this challenge, we propose a look-ahead computation and error compensation design that concurrently performs WAQ LUT-GEMM computations and outlier detection.
Fig.~\ref{fig:error_compensation} shows an example of the proposed look-ahead computation and error compensation design, which are performed in two parallel branches: the main branch (left) and the outlier branch (right).
In this $M-K-N$ GEMM example, $M = 1$, $K = 6$, $N = 4$, and $n_W = n_A = 1$.
There are two activation outliers in channels 2 and 4, indicated by deep green boxes.

\subsubsection{Look-Ahead Computations}
In the main branch, the entire FP16 activation vector is clustered to the nearest centroids in the activation codebook $\mathcal{C}_A$, producing quantized activations.
While the quantization error of the inliers is generally small and does not lead to significant accuracy degradation, the quantization error of the outliers is non-negligible.
For instance, the activation outliers in channels 2 and 4, originally $5.07$ and $-3.01$, are quantized to $0.78$ and $0.13$, yielding quantization errors (residuals) of $4.29$ and $-3.14$, respectively.
Our proposed look-ahead computation design allows the main branch to temporarily ignore the quantization errors of outliers and directly perform WAQ LUT-GEMM computations using the quantized activations, while the outlier branch simultaneously computes the error compensation for the outliers.

\subsubsection{Error Compensation}
In the outlier branch, the outlier detection units dynamically identify the outliers in the activation vector and compute their quantization errors by subtracting the quantized activation outliers from the original FP16 activation outliers.
We use the channel index of the activation outliers to fetch the corresponding input channels of quantized weights from the $idx_W$ matrix, which are then dequantized to FP16 format based on the weight codebook $\mathcal{C}_W$.
Then, the residuals are multiplied by the dequantized weights to generate error compensation terms, which are accumulated into the look-ahead computation results using WAQ LUT-GEMM from the main branch.
As shown in Fig.~\ref{fig:look_ahead}, this mechanism removes the time-consuming dynamic outlier detection from the critical path, enabling reduced GEMM runtime and ensuring identical mathematical results.

Note that in each cycle, the outlier detection unit sequentially outputs each pair of outlier value and their channel index.
Thus, in each cycle, only one input channel of the weight index matrix is fetched and dequantized for error compensation.
Compared to the conventional design for dynamic outlier identification (Fig.~\ref{fig:look_ahead}(a)) which performs sparse FP16 GEMMs for all residuals simultaneously, this design eliminates the need for sparse representation of the outliers and significantly reduces the number of multiply-accumulate (MAC) units required in the outlier branch.

%% file: 4_hardware.tex
\section{Hardware Design} \label{sec:hardware}
To enable the efficient execution of the computation optimizations introduced in \S\ref{sec:algorithm}, in this section, we present the supporting accelerator design of {\name}.
First, we present the overall architecture of the {\name} accelerator.
Then, we describe the micro-architecture designs of the Index Counter and the Clustering Unit, which are the key hardware components to support the proposed WAQ LUT-based GEMM.
Moreover, we also propose a lightweight outlier detection engine \textit{Orizuru}, which dynamically identifies the top-$k$ largest and smallest elements in each activation token with minimal overhead.

\subsection{Overall Architecture} \label{sec:arch-overview}

\begin{figure}
    \centering
    \includegraphics[width=\linewidth]{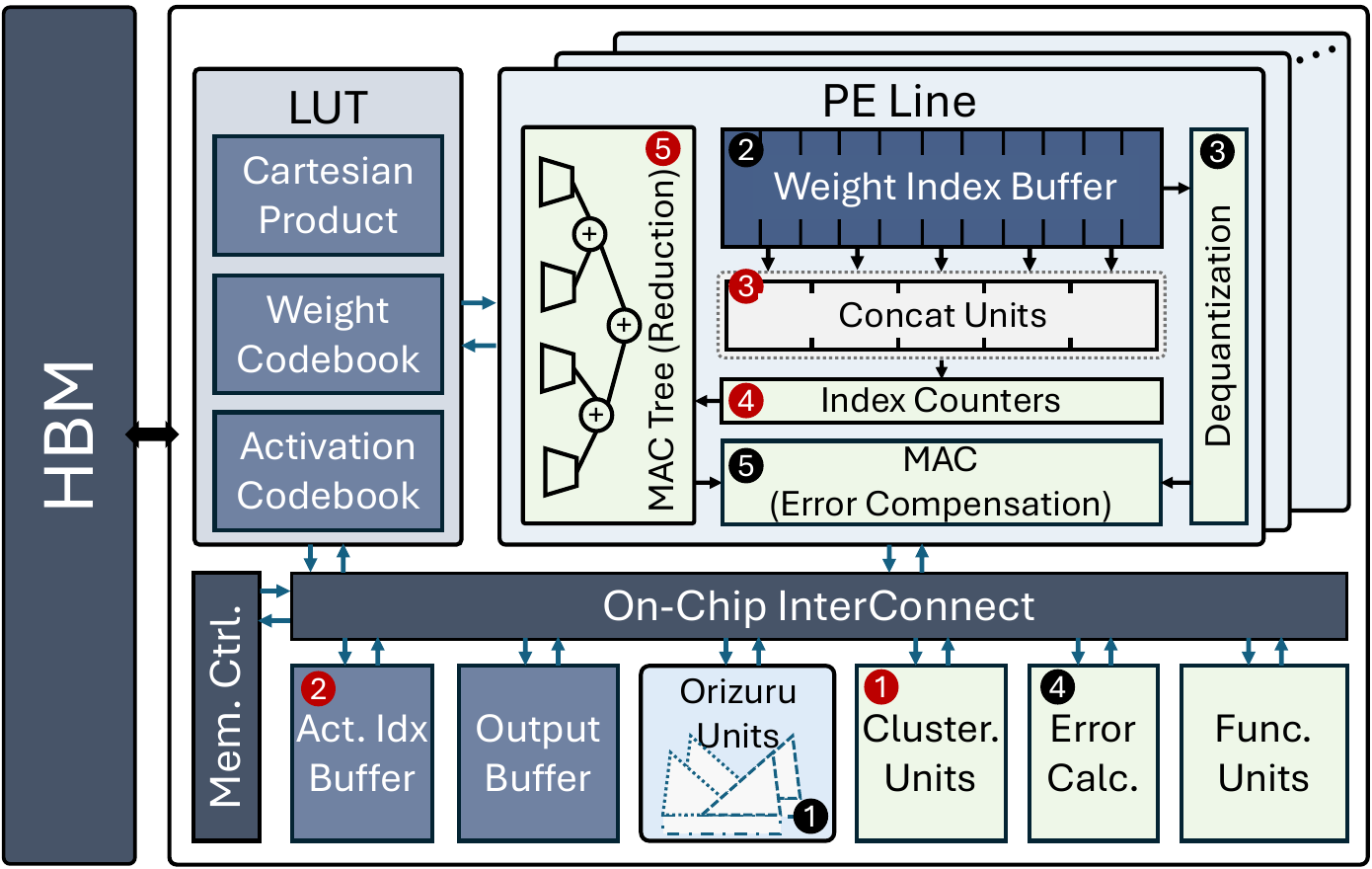}
    \vspace{-0.3in}
    \caption{Overall architecture of the {\name} accelerator.}
    \vspace{-0.2in}
    \label{fig:kllm-arch}
\end{figure}
Fig.~\ref{fig:kllm-arch} shows the overall architecture of the {\name} accelerator, and Table~\ref{tabl:hw-config} presents its hardware configurations.
The {\name} accelerator is composed of 16 Processing Element (PE) Lines, 
a LUT, an Activation Index Buffer, an Output Buffer, \textit{Orizuru} units, Clustering Units, Functional Units, an Error Calculation Unit, and a Memory Controller.
Besides the Cartesian Product results, the LUT also stores the centroid codebooks of activations and weights for the purposes of activation clustering and weight dequantization, respectively.
The main compute unit within the PE Lines is the Concat Unit, which accepts two 4-bit indices, concatenates them, and stores the result in an 8-bit register.
The Concat Unit's minimalist design provides high area efficiency compared to FP16 MAC units, benefiting both compute-intensive and memory-intensive scenarios.
For compute-intensive workloads such as LLM prefill phase, this efficiency enables a higher number of Concat Units on the chip, delivering the computational parallelism needed for high throughput.
For memory-intensive workloads such as LLM decode phase, the lightweight Concat Units consume minimal chip area, freeing up chip area for additional I/O pins to increase bandwidth and mitigate memory bottlenecks.

Within each PE Line, there are 4096 Concat Units, a Weight Index Buffer, 32 Index Counters, a Dequantization Unit, a MAC Tree for reduction, and 8 MAC Units for error compensation.
The computation flow of the main and outlier branches is illustrated with the red and black circled numbers, respectively.

For the main branch, in step \circlednumberred{1}, the Clustering Units reads the entire activation vector in the Output buffer, and clusters them to the nearest centroids based on the Activation Codebook.
The clustered activation indices are then stored in the Activation Index Buffer.
In step \circlednumberred{2}, the clustered activation indices are fetched from the Activation Index Buffer and broadcast to all PE Lines.
In step \circlednumberred{3}, the Concat Units in each PE Line concatenate the activation indices with the weight indices fetched from the Weight Index Buffer to form the concatenated indices.
Next, in step \circlednumberred{4}, the Index Counters calculate the counts of the unique concatenated indices.
Finally, in step \circlednumberred{5}, the MAC Tree performs the weighted-sum operations based on the counts and the corresponding Cartesian Product values retrieved from the LUT, generating the look-ahead computation results.

For the outlier branch, in step \circlednumberblack{1}, the \textit{Orizuru} units read the FP16 activations in the Output buffer and dynamically identify the outliers, sequentially outputting each outlier value along with its channel index.
The weight input channel corresponding to the outlier channel index is fetched from the Weight Index Buffer in step \circlednumberblack{2}, and dequantized to FP16 weights in the Dequantization Unit in step \circlednumberblack{3}.
In step \circlednumberblack{4}, the Error Calculation Unit calculates the residual between the outlier activation and its nearest centroid from the Activation Codebook.
Step \circlednumberblack{4} is performed in parallel with \circlednumberblack{2} and \circlednumberblack{3}.
Then, in step \circlednumberblack{5}, the MAC Units perform multiply-accumulates between the outlier residuals and the dequantized weights, generating the error compensation terms.
Finally, the MAC units combine the error compensation terms with the look-ahead computation results from the main branch to produce the final output activations, which are stored back in the Output Buffer.

To minimize end-to-end latency, the Memory Controller orchestrates pipelined execution of operations across both the main and outlier branches.
The cycle latencies for each computation step are presented in \S~\ref{sec:cycle_latency}.

\begin{figure}[t]
    \centering
    \includegraphics[width=\linewidth]{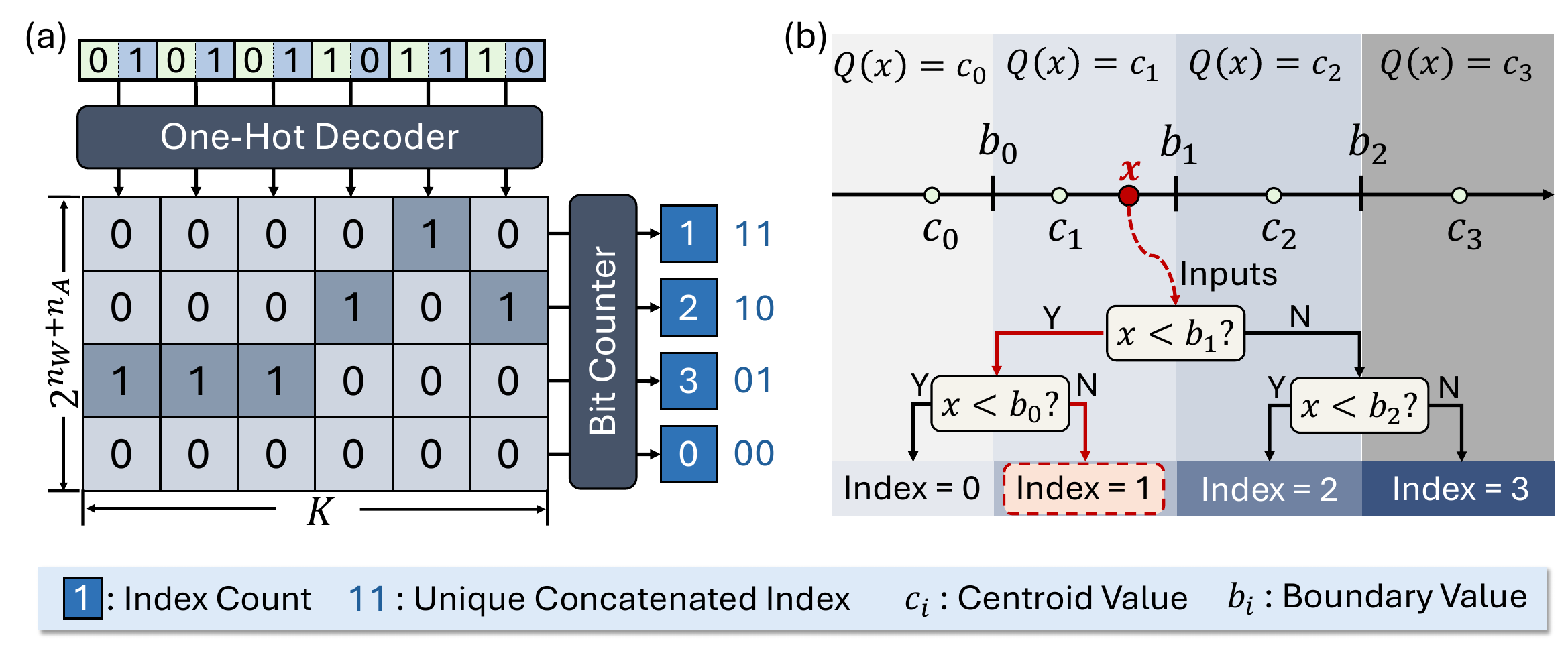}
    \vspace{-0.3in}
    \caption{Design of (a) the Index Counter and (b) the Clustering Unit.}
    \vspace{-0.2in}
    \label{fig:idx_counter}
\end{figure}

\begin{table}
\centering
\caption{{\name} accelerator configurations (28nm, 500MHz).}
\vspace{-0.1in}
\label{tab:split_module_subcolumns}
\resizebox{\columnwidth}{!}{
\begin{tabular}{c|c|c|c|c}
\toprule
\multicolumn{2}{c|}{\textbf{Module}} & \textbf{Specification} & \textbf{Area} \textit{$(mm^2)$} & \textbf{Power} \textit{$(W)$} \\ 
\midrule
\multicolumn{2}{c|}{PE Line}  &  16 PE Lines per chip   & 9.08 & 7.54 \\ 
\midrule
\multirow{5}{*}{PE Line} & Concat Unit   & 4096 Concat Units per line  & $8.68\times 10^{-2}$ & $8.36\times 10^{-2}$\\ 
                         & Wgt Idx Buffer    & {2}\thinspace{KB} per line   & $6.75\times 10^{-2}$ & $1.69\times 10^{-2}$ \\ 
                         & Index Counter    & 32 16-in Index Counters per line       & $2.71\times 10^{-1}$ & $6.14\times 10^{-2}$ \\ 
                         & Dequant Unit   & 1 Dequant Unit per line       & $2.83\times 10^{-3}$ & $6.11\times 10^{-3}$ \\ 
                         & MAC Tree & 1 32-in FP16 MAC Tree per line & $1.17\times 10^{-1}$ &$ 2.54\times 10^{-1}$ \\
                         & MAC           & 8 FP16 MAC Units per line & $2.26\times 10^{-2}$ &$ 4.89\times 10^{-2}$ \\ 
\midrule
\multicolumn{2}{c|}{Output Buffer}   & {64}\thinspace{KB} per chip        & 2.17 & $2.68\times 10^{-1}$\\ 
\midrule
\multicolumn{2}{c|}{Act. Idx Buffer}    & {16}\thinspace{KB} per chip         & $5.40\times 10^{-1}$  & $6.71\times 10^{-2}$ \\ 
\midrule
\multicolumn{2}{c|}{LUT} & {2}\thinspace{KB} per chip         & $6.75\times 10^{-2}$ & $8.38\times 10^{-3}$ \\ 
\midrule
\multicolumn{2}{c|}{Cluster. Unit	}   & 4 Clustering Units per chip & $1.31\times 10^{-3}$ & $2.90\times 10^{-4}$\\ 
\midrule
\multicolumn{2}{c|}{Orizuru	}    & 273 16-in Orizuru Units per chip & $7.39\times 10^{-1}$ & $2.73\times 10^{-1}$ \\ 
\midrule
\multicolumn{2}{c|}{Error Calc. Unit}   &1 Error Calculation Unit per chip & $4.12\times 10^{-3}$ & $6.40\times 10^{-3}$ \\ 
\midrule
\multicolumn{2}{c|}{Func. Unit	}   &1 Functional Unit per chip & $8.89\times 10^{-1}$ & $5.63\times 10^{-1}$ \\ 
\midrule
\multicolumn{2}{c|}{Memory Controller}   & 1 Memory Controller per chip & $1.47$ & $9.28\times 10^{-1}$ \\
\midrule
\multicolumn{2}{c|}{Total}   &---       & 15.31 & 9.66 \\ 
        
\bottomrule
\end{tabular}
}
\label{tabl:hw-config}
\end{table}

\subsection{Index Counter} \label{sec:idx-counter}

In {\name}, we design an Index Counter to compute the index distribution of the concatenated indices with high parallelism, which is illustrated in Fig.~\ref{fig:idx_counter}(a).
The concatenated indices are decoded into one-hot vectors in parallel, and the bit counters calculate the row-wise sums of the one-hot vectors to obtain the count for a certain concatenated index.
For example, the first concatenated index `01' is decoded into the one-hot vector [0, 0, 1, 0] as shown in the first column of the decoded one-hot matrix in Fig.~\ref{fig:idx_counter}(a).
In the first row of the decoded one-hot matrix, there is one `1', indicating that the concatenated index `11' appears once in the concatenated indices.
The row-wise sums of the second, third, and fourth rows correspond to the counts of concatenated indices `10', `01', and `00', respectively.
To satisfy timing and area constraints, the Index Counter employs a 16-input design, with each PE Line incorporating 32 Index Counters to perform the index distribution calculations in parallel.

\subsection{Clustering Unit} \label{sec:clustering_unit}
To perform efficient non-uniform quantization for the activations, we design a Clustering Unit as illustrated in Fig.~\ref{fig:idx_counter}(b), which maps each activation to its nearest centroid based on the pre-defined codebook.
The clustering unit first computes the boundary values between each pair of adjacent centroids: $b_i = (c_i + c_{i+1}) / 2$, where $c_i$ and $c_{i+1}$ are two adjacent centroids.
For any input value which is within $[b_{i-1}, b_i)$, it is assigned to the $i$-th cluster with centroid $c_i$.
Then, for each input activation $x$, the clustering unit compares it with the boundary values to determine the cluster index it belongs to.
To accelerate the clustering process, we implement the comparison logic using a binary search tree structure.
In the example of Fig.~\ref{fig:idx_counter}(b), there are 4 centroids, and each input activation $x$ undergoes $log_2(4) = 2$ hierarchical comparisons to determine its cluster index.

\subsection{Orizuru: Dynamic Outlier Detection Engine} \label{sec:orizuru}

\begin{figure*}[t!]
    \centering
    \includegraphics[width=\linewidth]{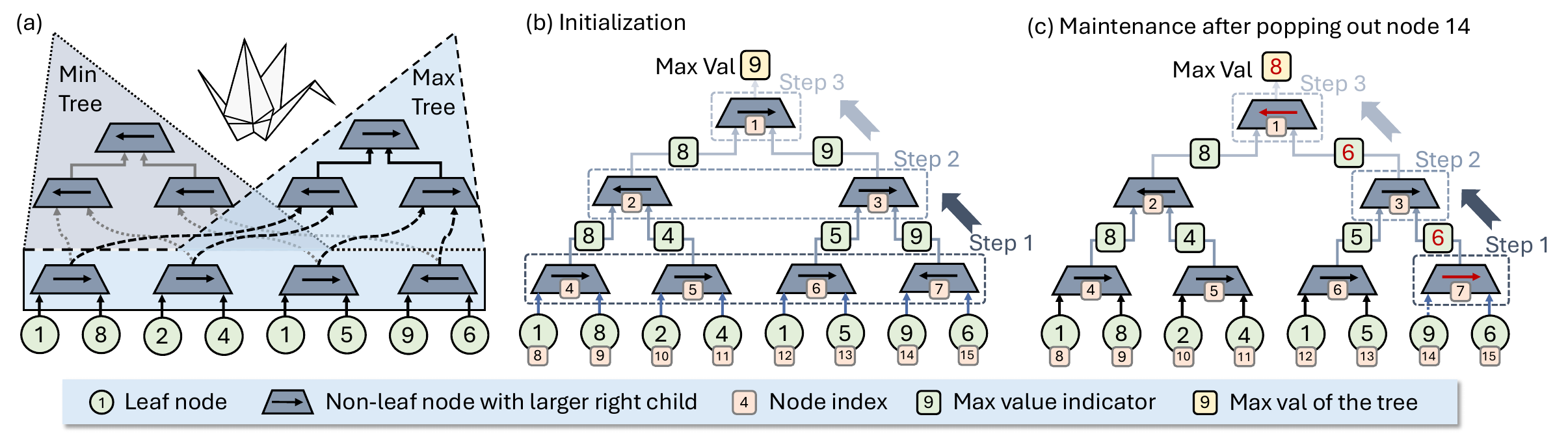}
    \vspace{-0.3in}
    \caption{Orizuru architecture. (a) The overall architecture of Orizuru: two-fold binary trees with shared leaf nodes and comparison results at the last layer of the non-leaf nodes. (b) Initialization of the max tree. (c) Maintenance of the max tree after popping out node 14.}
    \vspace{-0.2in}
    \label{fig:orizuru}
\end{figure*}

As mentioned in \S~\ref{sec:look_ahead}, {\name} minimizes runtime by performing outlier detection on a dedicated outlier branch that runs in parallel with the main branch.
To avoid making the outlier branch the performance bottleneck of the entire GEMM operation, we design \textit{Orizuru}
\footnote{The shape of the two-fold binary trees with shared leaves resembles an \textit{Orizuru} (a paper crane).}, an efficient outlier detection engine that identifies the outliers from each activation token with minimal latency and energy consumption.

We propose to detect the top $k$ largest and smallest values within each activation token $\mathbf{x} = [x_i] \in \mathbb{R}^{N}$, which are excluded from the quantization process.
To minimize the number of comparisons required for outlier detection, we adopt a tree-based architecture which maximizes the reuse of comparison results.
As shown in Fig.~\ref{fig:orizuru}(a), \textit{Orizuru} is composed of two complete binary trees with shared leaf nodes. 
These two binary trees, named the max tree, $\mathcal{P}$, and the min tree, $\mathcal{Q}$, are used to pop the $k$ largest and smallest elements of a given vector $\mathbf{x}$, respectively.
The tree structure offers high data reuse for input data, reducing memory access overhead. 
Without loss of generality, we take the max tree as an illustration of how we initialize, pop out, and maintain the tree; then we showcase how we reuse the information from the max tree to the min tree towards the \textit{Orizuru} architecture.

\myparatight{Complete binary tree architecture}
For simplicity, in the example of Fig.~\ref{fig:orizuru}, we consider an 8-input \textit{Orizuru} for $N=8$.
Specifically for the max tree, $\mathcal{P}$, as shown in Fig.~\ref{fig:orizuru}(b), we build a complete binary tree with $L$ levels, where $L = \log_2(N)$. 
Each node on the tree is a 2-to-1 multiplexer (MUX) controlled by the value in a bit buffer, denoted as $p_{l, i} \in [0, 1]$, where $l=1, 2, \dots, L$ is the level index and $i = 1, 2, \dots, 2^{l-1}$ is the node index of this level.
For the $N/2$ leaf nodes on the $L$-th level, the node $p_{L, i}$ is directly connected to the FP16 values of $x_{2i-1}$ and $x_{2i}$; the non-leaf node $p_{l, i}$ is connected to its left/right children $p_{l+1, 2i-1}$ and $p_{l+1, 2i}$, and its MUX outputs 
\looseness=-1
\begin{align}
    \text{MUX}(p_{l, i}) = 
    \begin{cases}
        \text{MUX}(p_{l+1, 2i-1}) ~\textrm{or}~ x_{2i-1},  & p_{l, i} = 0 \\
        \text{MUX}(p_{l+1, 2i}) ~\textrm{or}~ x_{2i},  & p_{l, i} = 1.
    \end{cases}
\end{align}

In the max tree, $\mathcal{P}$, the bit buffer points to the larger child node or activation value, which represents the largest element in the sub-tree rooted at the current node. 
Consequently, the output of the root node, $\textsf{MUX}(1, 1)$, corresponds to the largest value in the entire activation vector, $\max(\mathbf{x})$.
To track the availability of activation elements, we introduce a mask vector $\mathbf{m}^{(p)} = [m_i^{(p)}] \in [0, 1]^{N}$ for the max tree $\mathcal{P}$, where $m_i^{(p)} = 1$ indicates that the activation element $x_i$ is available, and $m_i^{(p)} = 0$ indicates it has already been popped out. 
Similarly, another independent mask vector $\mathbf{m}^{(q)} = [m_i^{(q)}] \in [0, 1]^{N}$ is defined for the min tree $\mathcal{Q}$.

\myparatight{Initializing the binary tree}
To initialize the max and min trees for a new activation vector $\mathbf{x}$, we update all the bit buffers tree, which is shown in Fig.~\ref{fig:orizuru}(b) for the max tree.
This process is performed in a bottom-up manner across the $L$ levels of the tree.
Initially, at the $\log_2(N)$-th level, the registers of the non-leaf nodes (e.g., nodes 4, 5, 6, and 7) are updated by comparing their left and right child nodes at the leaf level, which hold the input values.
This step involves $N / 2$ comparisons.
After completing these comparisons, the registers are updated to indicate whether the left or right child contains the larger value.
Next, this comparison process is repeated recursively for each level, moving upwards until the root node's register is updated.
Finally, the MUX at the root node selects the maximum value from the $N$ elements.
Overall, the tree initialization requires $N-1$ FP16 comparisons.

\myparatight{Popping out and maintaining the tree}
Once the initialization is complete, the maximum value can be popped out immediately. 
The index of this maximum value is determined by traversing the tree from the root node to the corresponding leaf node, following the binary digits stored in the registers. 
For example, to locate the index of the maximum element ``9'' (at node 14) in Fig.~\ref{fig:orizuru}(b), we start at the root node, which contains ``1'', directing us to its right child, node 3. 
Node 3 also contains ``1'', further directing us to its right child, node 7. 
This process continues for $\log_2(N)$ steps until we reach the maximum element ``9'' at node 14. 
The binary digits from the accessed registers are concatenated to form ``110''. 
Since leaf nodes are indexed from $N$ to $2N-1$, a ``1'' is prefixed to the concatenated value, resulting in ``1110''—the binary representation of the index ``14'' for the maximum value ``9''.
Note that this process does not involve any comparisons and can be completed in one clock cycle.

To identify the second largest value, the popped maximum value must first be effectively ``removed'' from the tree, which is the tree maintenance process.
This is achieved by updating the registers of the non-leaf nodes while maintaining the tree structure.
As shown in Fig.~\ref{fig:orizuru}(c), the updated information is highlighted with red Max val indicators and arrows.
Specifically, the popped maximum value is treated as negative infinity during the update for the register value of its parent node, while the original value remains unchanged in the FP16 buffer of the corresponding leaf node.
Consequently, the arrow in the parent node flips to point to the neighbor node of the popped value.
\looseness=-1

The registers of the ancestors of the popped element are updated in a bottom-up manner.  
This maintenance process resembles the initialization but involves only a single comparison per level.  
Each maintenance step requires $ \log_2(N)$ sequential comparisons.  
To retrieve the top-$k$ maximum values, this popping and maintenance process is repeated $k$ times.

Note that after repeating the maintenance process multiple times, there may be cases where both leaf nodes under a certain non-leaf node are popped out. 
For instance, after retrieving the top-3 largest values, both child nodes of node 7 may be popped out. 
In this case, the MUX of node 7 outputs negative infinity for the register update on its parent node (node 3) in the subsequent cycle. 
Obviously, as long as the tree is not entirely emptied, i.e., fewer than $N$ elements have been popped out, this negative infinity cannot propagate all the way to the root node.

\myparatight{Combining two trees into \textit{Orizuru}}
A key advantage of \textit{Orizuru}'s two-fold binary tree architecture is its ability to reuse the max tree's results to reduce the number of comparisons needed for the min tree.  
Given the limited number of comparators, the runtime bottleneck in \textit{Orizuru} occurs during the initial tree setup, which requires $N/2$ comparisons.  
However, the comparison results from this step can be directly reused to initialize the min tree.  
Specifically, the registers at the $\log_2(N)$-th level of the min tree can adopt the reversed comparison results from the max tree.  
This allows the min tree's initialization to skip the $\log_2(N)$-th level and begin directly at the $\left(\log_2{N}-1\right)$-th level, reducing the total comparisons for its initialization by 50\%.

In summary, \textit{Orizuru} picks the $k$ maximums and $k$ minimums from an $N$-input activation vector $\mathbf{x}$ at the cost of $1.5N + 2k \cdot \log_2(N)$ comparisons.
This is significantly smaller than the $6N$ comparisons required by the top-$k$ engine in~\cite{wang2021spatten}.

\myparatight{Dealing with ties}
Since the outlier detection is performed on the FP16 activations, which have limited precision, multiple activation values within each token can be identical, leading to ties when determining the $k$-th largest or smallest values.
On average, such ties occur in approximately 2\% of activation tokens across all evaluated layers and models.
To address this issue, we always output exactly $k$ outliers for each of the max and min trees to maintain a consistent number of outliers for error compensation.
This is achieved by the following tie-breaking strategy: in cases where there are ties in the comparison results, we deterministically select the left child node as the larger value in the max tree and as the smaller value in the min tree.

%% file: 5_evaluation.tex
\section{Evaluation}
\subsection{Experimental Setup}
\label{sec: exp setup}

\begin{table*}[t]
\centering
\caption{WikiText2 perplexity with 2048 sequence length.}
\vspace{-0.1in}
\label{tab:wikitext2_PPL}
\resizebox{\textwidth}{!}{
\begin{tabular}{l|c|ccc|ccc|ccc|c|c}
\toprule
\multirow{2}{*}[-1mm]{\textbf{Precision}} 
& \multirow{2}{*}[-1mm]{\textbf{Method}} 
& &\textbf{OPT}&& 
&\textbf{LLaMA}&&
&\textbf{LLaMA-2}&& 
\multirow{2}{*}[-1mm]{\textbf{LLaMA-3-8B}} 
& \multirow{2}{*}[-1mm]{\textbf{Mistral-7B}} \\

\cmidrule(lr){3-5} 
\cmidrule(lr){6-8} 
\cmidrule(lr){9-11}

 & & 6.7B & 13B & 30B 
   & 7B & 13B & 30B 
   & 7B & 13B & 70B 
   &  
   & \\

\midrule
\textbf{FP16} & - 
& 10.86 & 10.12 & 9.56 
& 5.68 & 5.09 & 4.10 
& 5.47 & 4.88 & 3.32 
& 6.14
& 5.25 \\

\midrule

\multirow{6}{*}{\textbf{W4A4}} 
& RTN 
& 6e3 & 3e4 & 7e3 
& 8e3 & 1e4 & 3e5 
& 2e3 & 7e3 & 2e5 
& 2e3 
& 6e3 \\

& SmoothQuant 
& 2e4 & 7e3 & 1e4 
& 4e2 & 67.20 & 32.51 
& 7e2 & 56.61 & 10.54 
& 1e3 
& 5e2 \\

& QuaRot 
& 12.21 & 11.20 & 10.92 
& 6.34 & 5.58 & 4.64 
& 6.19 & 5.45 & 3.83 
& 8.16 
& 5.77 \\

& $\text{Atom}^{\dagger}$ 
& 12.05 & 10.99 & 10.74 
& 6.25 & 5.52 & 4.61 
& 6.12 & 5.31 & 3.73 
& 8.10 
& 5.76 \\

& \cellcolor{gray!10}{\name}-S 
& \cellcolor{gray!10}{11.77} & \cellcolor{gray!10}{10.93} & \cellcolor{gray!10}{10.31} 
& \cellcolor{gray!10}6.08 & \cellcolor{gray!10}5.38 & \cellcolor{gray!10}4.40 
& \cellcolor{gray!10}6.00 & \cellcolor{gray!10}5.21 & \cellcolor{gray!10}3.60 
& \cellcolor{gray!10}{7.02} 
& \cellcolor{gray!10}5.84 \\

& \cellcolor{gray!20}{\namebf} 
& \cellcolor{gray!20}{\textbf{11.62}} & \cellcolor{gray!20}{\textbf{10.75}} & \cellcolor{gray!20}{\textbf{10.21}} 
& \cellcolor{gray!20}\textbf{6.04} & \cellcolor{gray!20}\textbf{5.37} & \cellcolor{gray!20}\textbf{4.38} 
& \cellcolor{gray!20}\textbf{5.90} & \cellcolor{gray!20}\textbf{5.19} & \cellcolor{gray!20}\textbf{3.55} 
& \cellcolor{gray!20}{\textbf{7.11}} 
& \cellcolor{gray!20}\textbf{5.75} \\

\midrule

\multirow{6}{*}{\textbf{W4A3}} 
& RTN 
& 3e4 & 2e4 & 2e4 
& 2e4 & 2e4 & 1e4 
& 6e5 & 5e5 & 6e5 
& 1e5 
& 1e4 \\

& SmoothQuant 
& 7e4 & 7e4 & 6e4 
& 5e4 & 2e4 & 2e4 
& 8e3 & 1e4 & 1e4 
& 8e3 
& 9e3 \\

& QuaRot 
& 2e2 & 2e2 & 1e2 
& 29.75 & 19.02 & 13.50 
& 2e2 & 2e2 & 85.28 
& 3e2 
& 2e2 \\

& $\text{Atom}^{\dagger}$ 
& 20.51 & 15.61 & 14.48 
& 9.62 & 7.36 & 6.18 
& 11.40 & 8.00 & 5.05 
& 13.11 
& 10.83 \\

& \cellcolor{gray!10}{\name}-S 
& \cellcolor{gray!10}{15.12} & \cellcolor{gray!10}{13.49} & \cellcolor{gray!10}{12.14} 
& \cellcolor{gray!10}7.60 & \cellcolor{gray!10}6.28 & \cellcolor{gray!10}5.31 
& \cellcolor{gray!10}7.91 & \cellcolor{gray!10}6.99 & \cellcolor{gray!10}4.13 
& \cellcolor{gray!10}{8.96} 
& \cellcolor{gray!10}7.42 \\

& \cellcolor{gray!20}{\namebf} 
& \cellcolor{gray!20}{\textbf{14.12}} & \cellcolor{gray!20}{\textbf{12.84}} & \cellcolor{gray!20}{\textbf{11.78}} 
& \cellcolor{gray!20}\textbf{7.17} & \cellcolor{gray!20}\textbf{6.21} & \cellcolor{gray!20}\textbf{5.10} 
& \cellcolor{gray!20}\textbf{7.49} & \cellcolor{gray!20}\textbf{6.43} & \cellcolor{gray!20}\textbf{4.05} 
& \cellcolor{gray!20}{\textbf{8.18}} 
& \cellcolor{gray!20}\textbf{7.27} \\

\bottomrule
\end{tabular}
}
\footnotesize{
\begin{tabular}{@{}l@{}}
\textit{$\dagger$} Atom applies group quantization to weights and activations, with the group size of 128. \\
\end{tabular}
}
\end{table*}

\myparatight{Models and Tasks}
We examine the algorithm performance of {\name} on a spectrum of LLMs and tasks.
Specifically, the models include OPT-6.7B/13B/30B~\cite{zhang2022opt}, LLaMA-7B/13B/30B~\cite{touvron2023llama}, LLaMA-2-7B/13B/70B~\cite{touvron2023llama2}, LLaMA-3-8B~\cite{grattafiori2024llama}, and Mistral-7B~\cite{jiang2023mistral}, which are implemented with Transformers~\cite{wolf2020transformers} and PyTorch~\cite{paszke2019pytorch}.
These LLMs are evaluated on two tasks:
(i) the next-word prediction task using the WikiText-2~\cite{merity2016pointer} dataset, measured by the perplexity (PPL) metric, and 
(ii) the zero-shot accuracy task across six common sense datasets: PIQA~\cite{bisk2020piqa}, ARC-easy (ARC-E)~\cite{clark2018think}, ARC-challenge (ARC-C)~\cite{clark2018think}, BoolQ~\cite{clark2019boolq},
HellaSwag~\cite{zellers2019hellaswag}, and WinoGrande~\cite{sakaguchi2021winogrande}.
The zero-shot performance evaluation utilizes the Language Model Evaluation Harness~\cite{eval-harness} framework.
All algorithm performance experiments are conducted on an NVIDIA A100-80GB GPU.

\myparatight{{\namebf}' NU-WAQ Implementation Details}
In {\name}, both weights and activations are quantized using K-Means clustering.
Weights undergo 4-bit per-output-channel quantization without outlier protection, while activations are quantized per-token with 3/4-bit precision.
We first perform post-training K-Means quantization on the LLM weights to obtain the weight centroids and indices.
Next, the activation centroids are trained offline using 16 calibration samples from the C4 dataset~\cite{dodge2021documenting}, and the indices are computed online.
We incorporate a weighted-K-Means algorithm to obtain the activation centroids, where the weights are determined by Fisher information matrices~\cite{pennington2018spectrum} of the activations.
To handle outliers, the top 0.5\% largest and bottom 0.5\% smallest activation values are preserved in FP16 format, while the inliers are quantized.
During inference, {\name} dynamically identifies outliers using the \textit{Orizuru} units, while {\name}-S reuses the thresholds from the offline training process on the calibration dataset.
\looseness=-1

\myparatight{Baseline LLM Quantization Methods}
We compare {\name} with INT-WAQ baselines round-to-nearest (RTN), SmoothQuant~\cite{xiao2023smoothquant}, QuaRot~\cite{ashkboos2024quarot}, and Atom~\cite{zhao2024atom}.
Except for Atom, which uses group-wise quantization for both weights and activations with a group size of 128, all other baseline algorithms employ per-output-channel quantization for weights and per-token quantization for activations.

\myparatight{Architecture Modeling and Comparison}
The hardware performance of the {\name} architecture is modeled using a cycle-accurate simulator modified from DnnWeaver~\cite{sharma2016high}.
The area and power metrics of the core logic units in the {\name} accelerator, such as the Concat Unit, MAC Tree, Index Counter, and \textit{Orizuru}, are derived from synthesis results using the TSMC {28}\thinspace{nm} standard cell library.
Table~\ref{tabl:hw-config} shows the detailed configurations of the hardware components on-chip.

We use Cacti~\cite{li2011cacti} and DRAMSim3~\cite{li2020dramsim3} to simulate the overhead of on-chip SRAM and off-chip HBM, respectively.
We denote W4A3 {\name} as {\name}-A3 and W4A4 {\name} as {\name}-A4, and similarly for {\name}-S.
We compare the hardware performance of {\name} with a series of baseline hardware accelerators, including the GPU-based platforms of NVIDIA A100-80GB GPU~\cite{nvidia-a100} and QuaRot~\cite{ashkboos2024quarot} and an ASIC accelerator of FIGLUT~\cite{park2025figlut}.
Unless otherwise specified, the hardware performance of {\name} and the baseline accelerators are evaluated on the next-word-prediction task with an output sequence length of 2048.

\begin{table*}
\centering
\caption{Zero-shot accuracy results with 2048 sequence length.}
\vspace{-0.1in}
\label{tab:zero_shot_accuracy}
\renewcommand{\arraystretch}{1.2}
\resizebox{0.8\linewidth}{!}{
\begin{tabular}{l|l|c|ccccccc}
\toprule
\multirow{2}{*}[-1mm]{\textbf{Model}} &
\multirow{2}{*}[-1mm]{\textbf{Precision}} & 
\multirow{2}{*}[-1mm]{\textbf{Method}} & 
\multicolumn{7}{c}{\textbf{Zero-Shot Accuracy} $\uparrow$} \\
\cmidrule(lr){4-10}
& & & PIQA & ARC-E & ARC-C & BoolQ & HellaSwag & WinoGrande & Avg. \\
\midrule

\multirow{9}{*}{\textbf{LLaMA-2-7B}}
& \textbf{FP16} & -- 
& 78.67 & 74.58 & 46.16 & 78.59 & 75.95 & 68.98 & 70.49 \\
\cline{2-10}

& \multirow{4}{*}{\textbf{W4A4}}
& QuaRot 
& 76.39 & 69.61 & 40.61 & 72.48 & 71.63 & 63.06 & 65.63 \\
& & $\text{Atom}^{\dagger}$ 
& 75.14 & 52.99 & 38.40 & 74.59 & 69.37 & 62.75 & 62.21 \\

& & \cellcolor{gray!10}\name-S
& \cellcolor{gray!10}77.31 & \cellcolor{gray!10}71.46
& \cellcolor{gray!10}42.92 & \cellcolor{gray!10}76.06
& \cellcolor{gray!10}72.57
& \cellcolor{gray!10}64.80 & \cellcolor{gray!10}67.52 \\

& & \cellcolor{gray!20}\namebf
& \cellcolor{gray!20}\textbf{77.97} & \cellcolor{gray!20}\textbf{73.06}
& \cellcolor{gray!20}\textbf{43.60} & \cellcolor{gray!20}\textbf{76.83}
& \cellcolor{gray!20}\textbf{74.32}
& \cellcolor{gray!20}\textbf{65.51} & \cellcolor{gray!20}\textbf{68.55} \\

\cline{2-10}

& \multirow{4}{*}{\textbf{W4A3}}
& QuaRot 
& 53.16 & 27.99 & 25.26 & 41.10 & 28.75 & 49.49 & 37.63 \\
& & $\text{Atom}^{\dagger}$ 
& 71.01 & 48.63 & 33.49 & 58.73 & 62.54 & 59.50 & 55.65 \\

& & \cellcolor{gray!10}\name-S
& \cellcolor{gray!10}75.14 & \cellcolor{gray!10}63.93
& \cellcolor{gray!10}37.37 & \cellcolor{gray!10}63.89
& \cellcolor{gray!10}67.58
& \cellcolor{gray!10}63.93 & \cellcolor{gray!10}61.97 \\

& & \cellcolor{gray!20}\namebf
& \cellcolor{gray!20}\textbf{75.84} & \cellcolor{gray!20}\textbf{65.99}
& \cellcolor{gray!20}\textbf{39.59} & \cellcolor{gray!20}\textbf{65.47}
& \cellcolor{gray!20}\textbf{68.28}
& \cellcolor{gray!20}\textbf{64.17} & \cellcolor{gray!20}\textbf{63.22} \\
\midrule

\multirow{9}{*}{\textbf{LLaMA-3-8B}}
& \textbf{FP16} & -- 
& 80.63 & 77.62 & 57.71 & 81.28
& 79.61 & 73.70 & 75.09 \\
\cline{2-10}

& \multirow{4}{*}{\textbf{W4A4}}
& QuaRot 
& 68.28 & 60.48 & 37.46 & 66.57
& 61.73 & 63.06 & 59.60 \\
& & $\text{Atom}^{\dagger}$ 
& 69.45 & 63.26 & 40.12 & 67.67
& 69.75 & 61.13 & 61.90 \\

& & \cellcolor{gray!10}\name-S
& \cellcolor{gray!10}77.62 & \cellcolor{gray!10}73.95
& \cellcolor{gray!10}50.34 & \cellcolor{gray!10}78.67
& \cellcolor{gray!10}75.88
& \cellcolor{gray!10}70.56 & \cellcolor{gray!10}71.17 \\

& & \cellcolor{gray!20}\namebf
& \cellcolor{gray!20}\textbf{78.67} & \cellcolor{gray!20}\textbf{74.03}
& \cellcolor{gray!20}\textbf{51.37} & \cellcolor{gray!20}\textbf{80.02}
& \cellcolor{gray!20}\textbf{77.00}
& \cellcolor{gray!20}\textbf{71.27} & \cellcolor{gray!20}\textbf{72.06} \\

\cline{2-10}

& \multirow{4}{*}{\textbf{W4A3}}
& QuaRot 
& 49.84 & 26.18 & 25.60 & 43.82
& 26.09 & 50.20 & 36.95 \\
& & $\text{Atom}^{\dagger}$ 
& 72.86 & 51.06 & 40.52 & 61.19
& 67.78 & 60.87 & 59.05 \\

& & \cellcolor{gray!10}\name-S
& \cellcolor{gray!10}75.82 & \cellcolor{gray!10}70.22
& \cellcolor{gray!10}41.99 & \cellcolor{gray!10}74.01
& \cellcolor{gray!10}71.98
& \cellcolor{gray!10}65.03 & \cellcolor{gray!10}66.51 \\

& & \cellcolor{gray!20}\namebf
& \cellcolor{gray!20}\textbf{77.09} & \cellcolor{gray!20}\textbf{71.89}
& \cellcolor{gray!20}\textbf{45.65} & \cellcolor{gray!20}\textbf{75.64}
& \cellcolor{gray!20}\textbf{73.80}
& \cellcolor{gray!20}\textbf{66.22} & \cellcolor{gray!20}\textbf{68.38} \\

\midrule

\multirow{9}{*}{\textbf{Mistral}}
& \textbf{FP16} & -- 
& 82.54 & 79.42 & 54.18 & 77.39
& 81.18 & 75.22 & 74.99 \\
\cline{2-10}

& \multirow{4}{*}{\textbf{W4A4}}
& QuaRot 
& 80.19 & 70.97 & 41.06 & 73.00
& 72.88 & 72.34 & 68.41 \\
& & $\text{Atom}^{\dagger}$ 
& 80.71 & 68.63 & 52.39 & 74.55
& 77.52 & 72.03 & 70.97 \\

& & \cellcolor{gray!10}\name-S
& \cellcolor{gray!10}81.77 & \cellcolor{gray!10}77.26
& \cellcolor{gray!10}51.81 & \cellcolor{gray!10}74.20
& \cellcolor{gray!10}78.99
& \cellcolor{gray!10}72.97 & \cellcolor{gray!10}72.83 \\

& & \cellcolor{gray!20}\namebf
& \cellcolor{gray!20}\textbf{82.10} & \cellcolor{gray!20}\textbf{77.82}
& \cellcolor{gray!20}\textbf{53.24} & \cellcolor{gray!20}\textbf{75.77}
& \cellcolor{gray!20}\textbf{80.15}
& \cellcolor{gray!20}\textbf{73.40} & \cellcolor{gray!20}\textbf{73.80} \\

\cline{2-10}

& \multirow{4}{*}{\textbf{W4A3}}
& QuaRot 
& 53.16 & 27.99 & 25.26 & 41.10
& 25.68 & 48.78 & 36.99 \\
& & $\text{Atom}^{\dagger}$ 
& 73.69 & 54.06 & 37.98 & 68.07
& 73.24 & 63.73 & 61.80 \\

& & \cellcolor{gray!10}\name-S
& \cellcolor{gray!10}77.35 & \cellcolor{gray!10}74.01
& \cellcolor{gray!10}43.68 & \cellcolor{gray!10}70.85
& \cellcolor{gray!10}76.02
& \cellcolor{gray!10}66.93 & \cellcolor{gray!10}68.14 \\

& & \cellcolor{gray!20}\namebf
& \cellcolor{gray!20}\textbf{79.87} & \cellcolor{gray!20}\textbf{76.30}
& \cellcolor{gray!20}\textbf{49.32} & \cellcolor{gray!20}\textbf{73.03}
& \cellcolor{gray!20}\textbf{78.58}
& \cellcolor{gray!20}\textbf{70.56} & \cellcolor{gray!20}\textbf{71.28} \\

\bottomrule
\end{tabular}
}
\footnotesize{
\begin{tabular}{@{}l@{}}
\textit{$\dagger$} Atom applies group quantization to weights and activations, with the group size of 128. \\
\end{tabular}
\vspace{-0.1in}
}
\end{table*}

\subsection{Algorithm Performance Analysis} \label{sec:algorithm_results}

Table~\ref{tab:wikitext2_PPL} shows the WikiText-2 PPL results for {\name} and baseline INT-WAQ methods across various models with a sequence length of 2048. 
{\name} consistently achieves the lowest PPL for both W4A4 and W4A3 precisions, outperforming the INT-WAQ methods. 
This demonstrates the effectiveness of {\name}'s NU-WAQ and outlier protection methods in reducing quantization errors and enhancing model performance.
For LLaMA-2-7B at W4A4, {\name} achieves a PPL of 5.90, with only a 0.43 degradation from the FP16 model, which is 34\% lower than Atom's degradation. 
Additionally, {\name} reduces PPL by 0.05 at W4A4 and 0.27 at W4A3 compared to {\name}-S, highlighting the benefits of dynamic outlier detection. 
For the LLaMA-3-8B, which is known to be more quantization-sensitive, {\name} achieves a PPL of 7.11 at W4A4, which reduces the PPL degradation by 49\% compared to Atom.
We notice that for LLaMA-2-7B and 13B, W4A3 quantization yields higher PPL than their counterparts in the LLaMA-7B and 13B models, because the more extensively trained LLaMA-2 models are harder to post-training quantize at low precisions~\cite{kumar2024scaling}.

On average, {\name} introduces only a 2.05\% and 5.90\% accuracy drop at W4A4 and W4A3 precision levels, respectively, compared to the FP16 baseline, while significantly outperforming state-of-the-art INT-WAQ methods.
In the W4A4 setting, {\name} improves accuracy by 6.44\% and 6.92\% compared to Atom and QuaRot, respectively.
Under the W4A3 configuration, {\name} achieves accuracy improvements of 8.79\% over Atom and 30.44\% over QuaRot.

\subsection{Hardware Performance Analysis} \label{sec: hardware performance analysis}

\begin{figure}
    \centering
    \includegraphics[width=\linewidth]{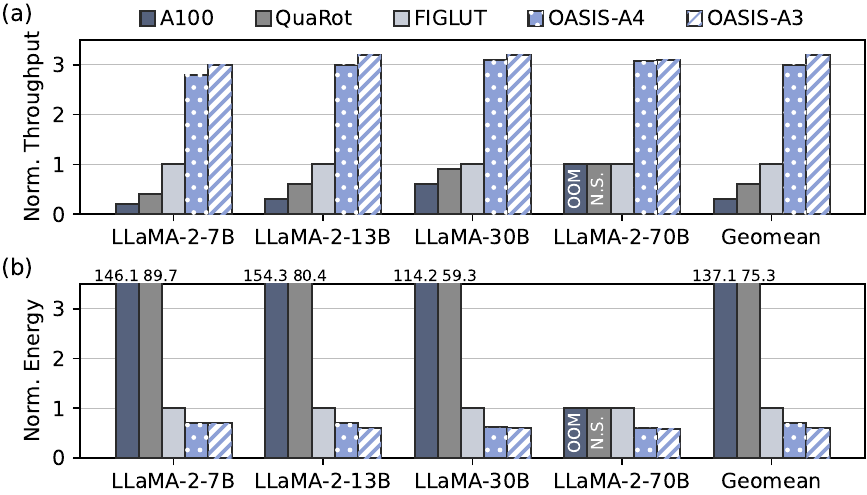}
    \vspace{-0.3in}
    \caption{Normalized throughput and energy consumption of {\name} and baseline accelerators in single-batch decoding.}
    \vspace{-0.1in}
\label{fig:diff-models}
\end{figure}

For hardware performance, we evaluate {\name} (an NU-WAQ design) against FP16, INT-WAQ, and WOQ LUT accelerators.
FP16 inference is run on the A100 GPU.
For the INT-WAQ baseline, we deploy QuaRot's W4A4 GEMM kernel on the A100, since Atom's kernel is only available for LLaMA-2-7B among the models we test (as reported in QServe~\cite{lin2024qserve}).
For the WOQ LUT comparison, we use FIGLUT, the SOTA ASIC LUT design evaluated at W4A16 precision.

Fig.~\ref{fig:diff-models} shows the normalized throughput and energy consumption of {\name} and baseline accelerators in single-batch decoding, with results normalized to FIGLUT.
N.S. indicates that the accelerator does not support the corresponding model, while OOM indicates that the accelerator runs out of memory for the specified model.
For throughput, on average, {\name}-A4 achieves 5.41$\times$, 3.12$\times$, 3.00$\times$ speedup and {\name}-A3 achieves 5.67$\times$, 3.27$\times$, 3.15$\times$ speedup over A100, QuaRot, and FIGLUT, respectively.
For energy efficiency, on average, {\name}-A4 achieves 198.1$\times$, 108.8$\times$, 1.44$\times$, and {\name}-A3 achieves 206.53$\times$, 113.56$\times$, 1.51$\times$ energy efficiency improvement over A100, QuaRot, and FIGLUT, respectively.
The performance of GPU-based accelerators (A100 and QuaRot) is limited by low batch sizes during single-batch decoding, while FIGLUT's performance is constrained by limited parallelism due to small group sizes.
In contrast, {\name} leverages an efficient WAQ LUT-GEMM design to substantially enhance computational parallelism, yielding superior throughput and energy efficiency.

\subsection{Ablation Studies} \label{sec:ablation}
\begin{figure}
    \centering

    \includegraphics[width=\linewidth]{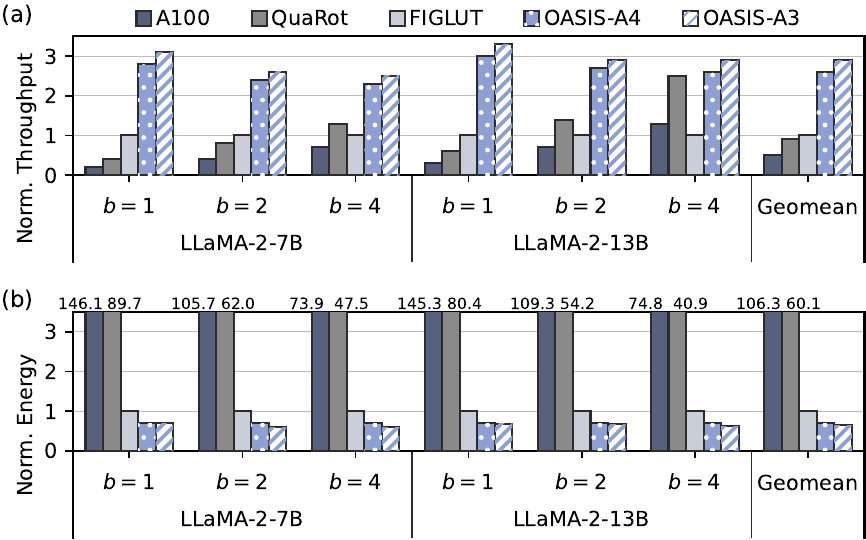}
    \vspace{-0.2in}
    \caption{Normalized throughput and energy consumption of {\name} and baseline accelerators during low-batch decoding.}
    \vspace{-0.1in}
    \label{fig:diff-batches}
\end{figure}

\subsubsection{Batched-Decoding}
{\name} is an ASIC accelerator targeting edge LLM inference, where low-batch decoding is the predominant use case.
In Fig.~\ref{fig:diff-batches}, we compare the normalized throughput and energy consumption of {\name}-A4/A3 over the baseline accelerators during low-batch decoding with batch sizes of 1, 2, and 4.
The evaluation is conducted with the LLaMA-2-7B/13B models.
{\name}-A4/A3 achieve average speedups of 3.41$\times$ and 3.73$\times$ over baseline accelerators, and average energy efficiency improvements of 26.43$\times$ and 28.20$\times$, respectively.
As the batch size increases, all accelerators exhibit higher throughput and lower energy consumption, primarily due to increased arithmetic intensity from weight reuse.
GPU-based approaches show steady throughput gains as batch size increases, which is because of higher Tensor Core utilization on GPUs~\cite{nvidia-tensor}. 
Nonetheless, {\name} still surpasses the baseline accelerators in both throughput and energy efficiency, especially with the smaller model of LLaMA-2-7B, which is more relevant for edge deployment.

\subsubsection{Prefill vs Decode}
\begin{figure}
    \centering
    \includegraphics[width=1\linewidth]{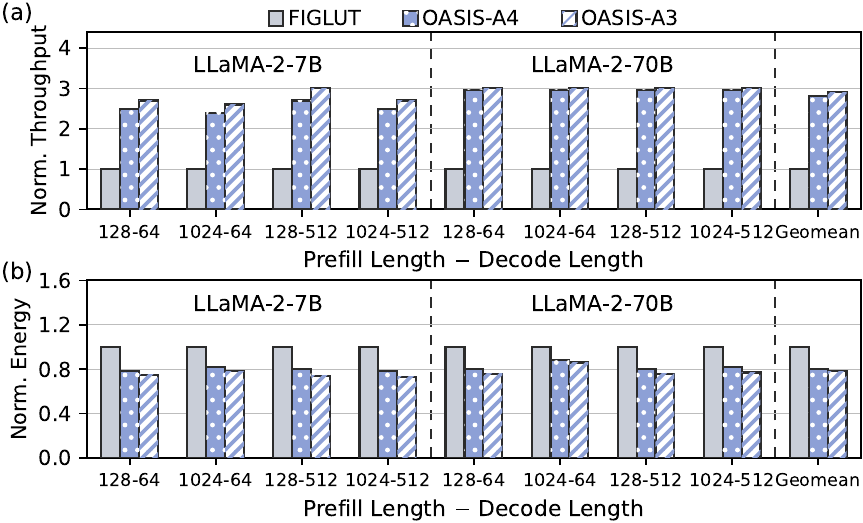}
    \vspace{-0.3in}
    \caption{Normalized throughput and energy consumption of {\name} and baseline accelerators for various prefill/decode length pairs.}
    \vspace{-0.1in}
    \label{fig:pd}
\end{figure}
We evaluate the performance of {\name} and FIGLUT under different prefill and decode length pairs using the LLaMA-2-7B/70B models, which is shown in Fig.~\ref{fig:pd}.
On average, {\name}-A4/A3 achieves 2.80$\times$ and 2.93$\times$ speedup over FIGLUT across different prefill/decode length pairs.
Notably, {\name}'s throughput and energy efficiency improvement over FIGLUT is more pronounced on the LLaMA-2-70B model than on the LLaMA-2-7B model, which is because larger models have a higher number of input channels, allowing {\name} to better leverage its compute efficiency advantage.

\subsubsection{Cycle Latencies for each Step in the Computation Pipeline} \label{sec:cycle_latency}
\begin{figure}[t]
    \centering
    \includegraphics[width=\linewidth]{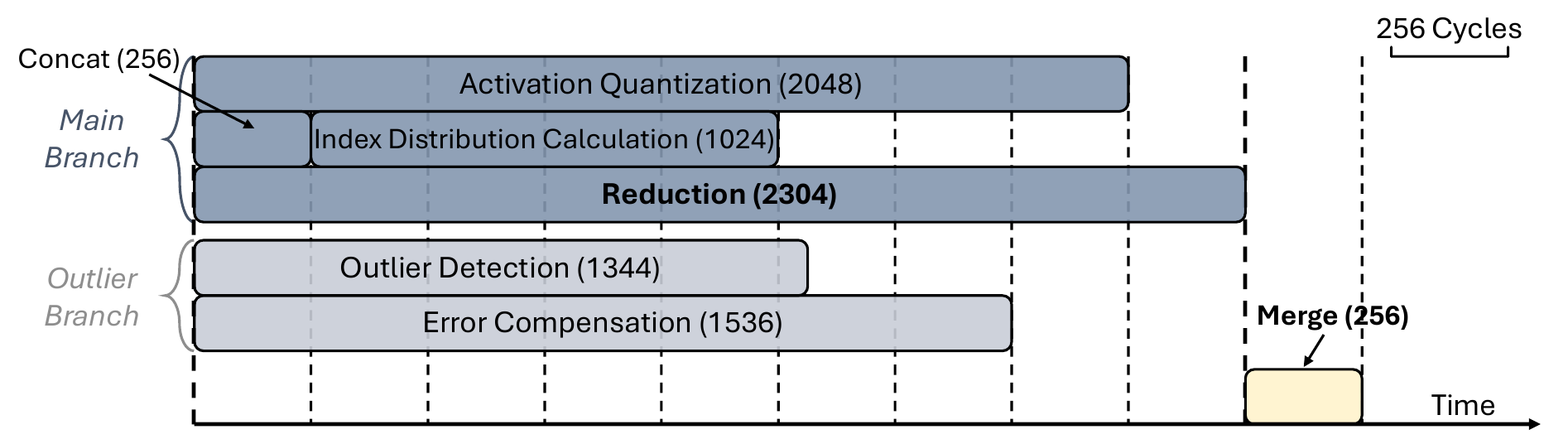}
    \vspace{-0.3in}
    \caption{Computation pipeline of performing an 1-4096-4096 GEMM with 1\% outliers on {\name} at W4A4 precision. The numbers in parentheses indicate the number of cycles required for each step. The steps that bottleneck each pipeline stage are bolded.}
    \vspace{-0.1in}
    \label{fig:cycle_latency}
\end{figure}

Fig.~\ref{fig:cycle_latency} shows the pipeline execution schedule of performing a 1-4096-4096 GEMM with {\name} at W4A4 precision with 1\% outliers.
The cycle latencies of each step are also shown in the figure with the numbers in parentheses.
Based on the hardware configurations in Table~\ref{tabl:hw-config}, in the 1\% outlier case, the two branches exhibit comparable latencies, with the outlier branch completing approximately 33\% faster.
Consequently, the outlier branch finishes first and outputs results to the Output Buffer, which are subsequently merged with the main branch results upon completion.
Conversely, in outlier-heavy scenarios, the main branch may finish first, with its results held in the Output Buffer awaiting the outlier branch completion.
\looseness=-1

\subsubsection{Outlier Sensitivity}

\begin{figure}
    \centering
    \includegraphics[width=\linewidth]{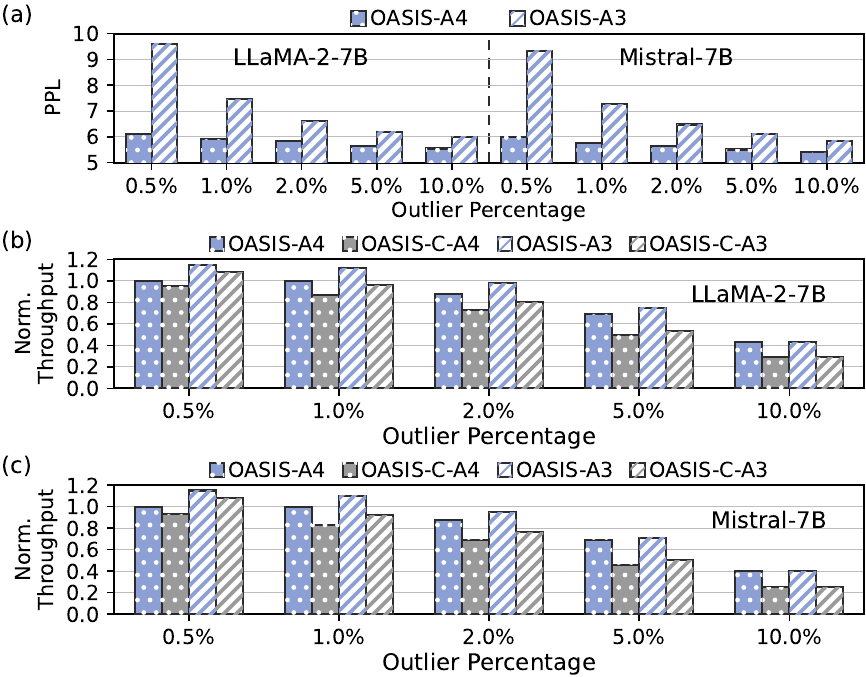}
    \vspace{-0.25in}
    \caption{(a) PPL, (b) LLaMA-2-7B's normalized throughput, and (c) Mistral-7B's normalized throughput of {\name} across different outlier percentages.}
    \label{fig:outlier_per_op_energy}
\end{figure}

Fig.~\ref{fig:outlier_per_op_energy}(a) presents the WikiText-2 PPL of LLaMA-2-7B and Mistral-7B on {\name} for outlier percentages ranging from 0.5\% to 10\%. For both models, increasing the outlier percentage generally improves PPL. To further examine the impact of increasing the outlier percentage on throughput, Fig.~\ref{fig:outlier_per_op_energy}(b) and (c) show the throughput of LLaMA-2-7B and Mistral-7B normalized to that of {\name}-A4, respectively. We make two observations: (i) increasing the outlier percentage from 0.5\% to 1\% results in negligible throughput degradation for both models, as the end-to-end latency is dominated by the main branch; (ii) further increasing the outlier percentage from 1\% to 10\% leads to a significant increase in the execution time of the outlier branch, which becomes the new bottleneck of the end-to-end latency. This is because, as discussed in \S~\ref{sec:arch-overview}, the hardware configurations in Table~\ref{tabl:hw-config} are chosen such that the execution times of the main and outlier branches are comparable at 1\% outlier percentage. Therefore, when the outlier percentage remains at or below 1\%, the outlier branch does not constitute a bottleneck; however, once it exceeds this threshold, the computational overhead of the outlier branch grows rapidly and dominates the overall latency.

To demonstrate the effectiveness of the look-ahead design, we quantify the latency of dynamic outlier detection by comparing {\name}'s throughput to the conventional dynamic detection design (Fig.~\ref{fig:look_ahead}(a), denoted as {\name}-C), where outlier detection is placed on the GEMM critical path.
On LLaMA-2-7B, when keeping 1\% of outliers, {\name}-A4 and {\name}-A3 achieve 16\% and 18\% higher throughput than {\name}-C-A4 and {\name}-C-A3, respectively, demonstrating the importance of the look-ahead design in hiding the latency of dynamic outlier detection and achieving high throughput.

\subsubsection{Comparisons with LUT-Based GEMM Designs}
\begin{figure}
    \centering
    \includegraphics[width=1\linewidth]{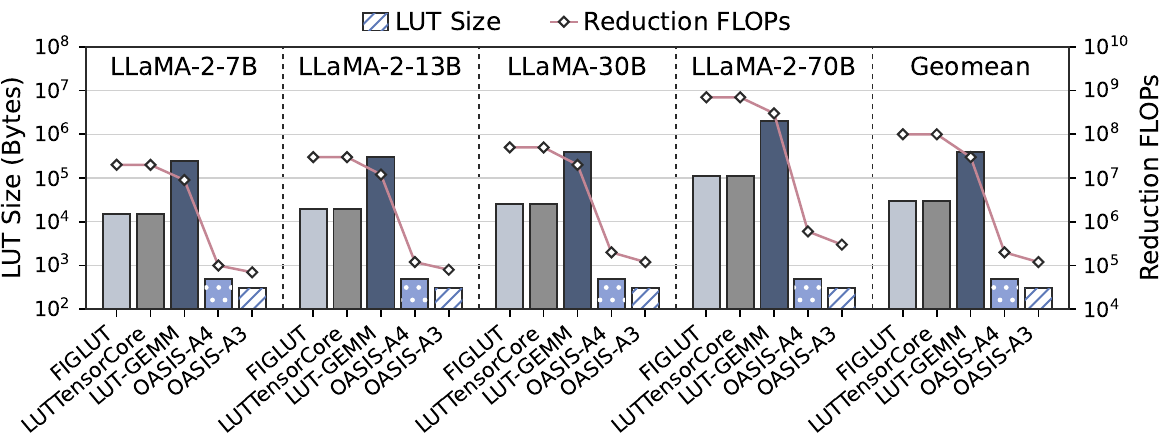}
    \vspace{-0.3in}
    \caption{LUT sizes and reduction FLOPs of {\name} and WOQ LUT-GEMM designs for the GEMM of the $q\_proj$ layer.}
    \vspace{-0.1in}
    \label{fig:lut_overhead}
\end{figure}

In Fig.~\ref{fig:lut_overhead}, we compare the LUT sizes and FLOPs during reduction of {\name} with WOQ LUT-GEMM designs, including FIGLUT~\cite{park2025figlut}, LUT Tensor Core~\cite{mo2025lut}, and LUT-GEMM~\cite{park2022lut}.
The evaluation is conducted on the $q\_proj$ layer's GEMM operation in different LLaMA models with W4A16 precision for WOQ LUT-GEMM designs.
On average, {\name}-A4 reduces LUT sizes by 62.1$\times$, and 994.2$\times$ compared to FIGLUT/LUT Tensor Core, and LUT-GEMM, respectively.
{\name}-A4 also decreases FLOPs during reduction by 497.1$\times$, and 248.6$\times$ compared to FIGLUT/LUT Tensor Core, and LUT-GEMM, respectively.
The three LUT baseline methods all employ Inner Product LUTs with small group sizes to limit LUT size.
This results in high FLOPs during reduction and consequently limits compute efficiency.
Among these, LUT-GEMM trades off LUT size for lower FLOPs during reduction by utilizing a larger group size.
As the model size increases from 7B to 70B, the number of input channels also increases from 4096 to 26728, leading to a significant rise in LUT sizes for all WOQ LUT-GEMM designs.
In contrast, {\name} adopts Cartesian Product LUTs, which enable constant LUT sizes regardless of the number of input channels.
As the model size increases, the increase of FLOPs in {\name} during reduction is also marginal compared to WOQ LUT-GEMM designs.

\subsubsection{Robustness of Offline-Learned Activation Centroids}
\begin{figure}[t]
    \centering
    \includegraphics[width=1\linewidth]{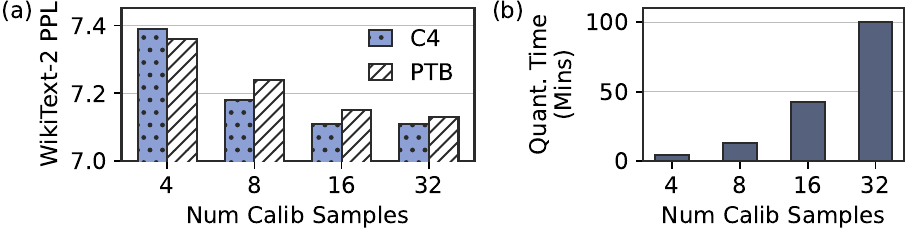}
    \vspace{-0.2in}
    \caption{Effects across calibration datasets and numbers of calibration samples on (a) PPL and (b) quantization time of {\name}-A4 on LLaMA-3-8B.}
    \label{fig:calib_dset}
\end{figure}

Fig.~\ref{fig:calib_dset} investigates how calibration dataset selection and sample quantity affect the PPL and quantization time of {\name}-A4 on LLaMA-3-8B.
As shown in Fig.~\ref{fig:calib_dset}(a), PPL remains consistent across different calibration datasets (C4 and PTB), with minimal variation.
For instance, at 16 samples, PPL is 7.11 (C4) versus 7.15 (PTB).
Generally, using C4 as the calibration dataset yielding slightly better PPL than PTB, which is because C4 is a larger and more comprehensive dataset than PTB, providing better coverage of the data distribution for centroid learning.
Increasing calibration samples from 4 to 32 improves PPL (7.39 → 7.11 for C4), but convergence occurs around 16 samples, beyond which quantization time grows substantially (42.47 → 100.52 minutes) without significant PPL gains.
Consequently, we employ 16 C4 samples for activation centroid learning in {\name} to achieve an optimal balance between accuracy and efficiency.

\subsubsection{Memory access / energy breakdown}
\begin{figure}
    \centering
    \includegraphics[width=\linewidth]{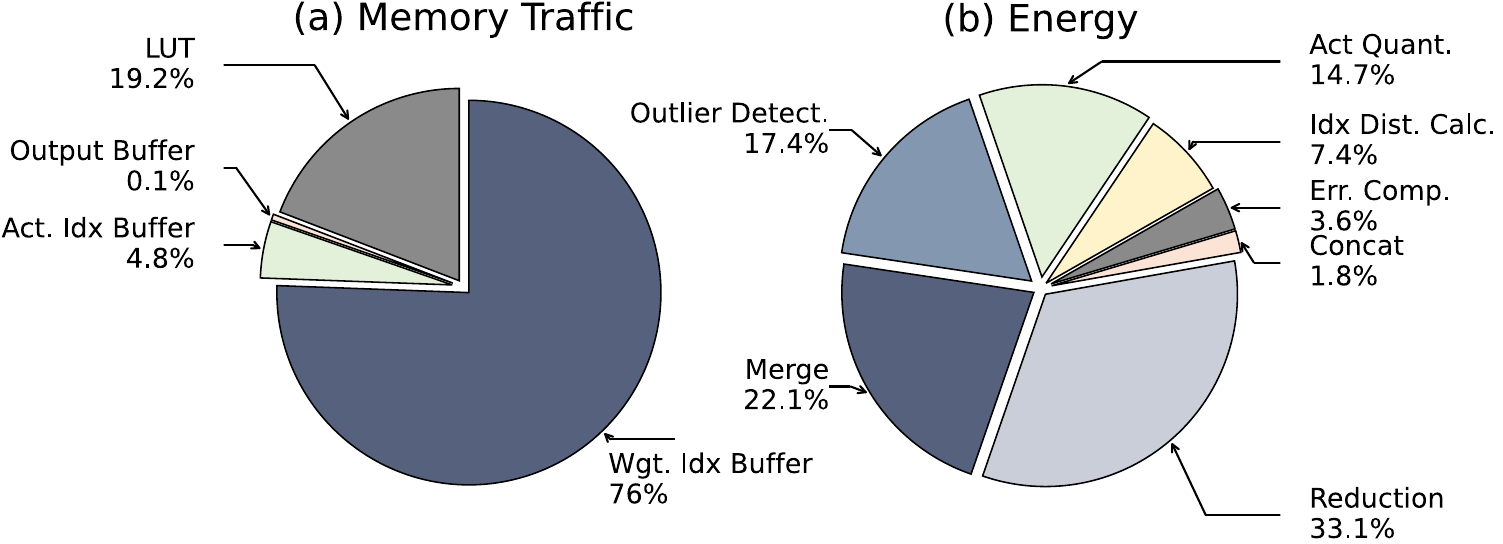}
    \vspace{-0.2in}
    \caption{Breakdown of (a) memory traffic and (b) energy consumption of {\name}-A4 for a 1-4096-4096 GEMM with 1\% outliers.}
    \vspace{-0.1in}
    \label{fig:mem_energy_breakdown}
\end{figure}

In Fig.~\ref{fig:mem_energy_breakdown}, we present the breakdown of on-chip memory traffic and energy consumption for a 1-4096-4096 GEMM with 1\% outliers with {\name}-A4.
Memory traffic is measured as the total number of bytes transferred, including both reads and writes.
The Weight Index Buffer dominates memory traffic at 76.0\%, while LUT reads and writes contribute 19.2\%, demonstrating that LUT access does not induce significant memory overhead.
Energy consumption is primarily attributed to reduction (33.1\%) and merging results from the main and outlier branches (22.1\%).

%% file: 6_relatedwork.tex
\section{Related Works}

\subsection{LLM WAQ Methods}
In WAQ settings, both weights and activations are quantized to low precision, which can significantly reduce memory usage and computational costs during LLM inference~\cite{zhao2024atom, liu2024spinquant, ashkboos2024quarot,gao2026disaggregated}.
For example, SmoothQuant~\cite{xiao2023smoothquant} applies scaling on both weights and activations to migrate the quantization difficulties of activations to weights, which are easier to quantize due to their smaller magnitude and quantity of outliers.
QuaRot~\cite{ashkboos2024quarot} applies Hadamard rotation matrices on both weights and activations to spread the quantization noise across all dimensions.
Atom~\cite{zhao2024atom} applies fine-grained quantization granularity to limit the impact of quantization noise caused by outliers within smaller groups, and preserve some outliers with higher precision.
However, they still lead to noticeable PPL degradation compared to FP16 models in low-precision configurations, and induce additional runtime overhead during GEMM operations.
In contrast, {\name} does not incorporate outlier suppression operations, and handle the outliers without additional runtime overhead.


\subsection{Reduction Tree-Based Architectures} \label{sec:tree_based}
Reduction trees perform O(N) operations with O(logN) latency by exploiting parallelism across tree levels. 
They are widely used for summation, e.g., in MAERI~\cite{kwon2018maeri} and Flexagon~\cite{munoz2023flexagon}, and can also support outlier selection via tournament trees~\cite{StepanovKershenbaum1986}, which identify maxima or minima through hierarchical pairwise comparisons.
Inspired by tournament trees, we develop \textit{Orizuru}, an outlier detection engine tailored for efficiently identifying both maximum and minimum activation outliers during LLM inference.
\textit{Orizuru} features shared leaf nodes between the maximum and minimum trees, which allows for efficient comparison of both maximum and minimum values with reduced hardware costs.

%% file: 7_conclusion.tex
\section{Conclusion} \label{sec:conclusion}
{\name} introduces an approach to executing NU-WAQ inference by eliminating dequantization and maximizing compute efficiency. 
By leveraging offline-computed Cartesian-product LUTs, {\name} significantly reduces LUT sizes and enables large-granularity GEMMs that exploit massive parallelism. 
Its outlier-aware quantization and lightweight \textit{Orizuru} top-$k$ engine further preserve accuracy and efficiency without adding runtime latency and with only marginal energy overhead.
Together, these innovations bridge the algorithm-hardware gap for NU-WAQ, maintaining accuracy with substantial throughput and energy efficiency gains.

\section*{Acknowledgement}

This work was partially supported by NSF under Grant Nos.~2332744, 2112562, and 2148253, and by AFOSR under Grant No.~FA9550-24-1-0322. 
The authors would like to thank Duke CEI Lab for their support. 
We also acknowledge helpful discussions with Yiran Chen, Zhixu Du and Changchun Zhou.